\RequirePackage{amsmath}
\documentclass{article}
\usepackage[utf8]{inputenc}

\usepackage{easyReview}
\renewcommand{\add}{}
\renewcommand{\highlight}{}

\usepackage{graphicx}
\usepackage{amsmath}
\usepackage{amsthm}

\usepackage{amssymb}
\usepackage{amsfonts}
\usepackage{booktabs}
\usepackage{algorithm}
\usepackage{algorithmic}
\usepackage{xcolor}
\usepackage{subfigure}
\usepackage{algorithm}
\usepackage{algorithmic}
\graphicspath{{./figures/}}
\usepackage{array}
\newcolumntype{C}[1]{>{\centering\let\newline\\\arraybackslash\hspace{0pt}}m{#1}}

\newtheorem{theorem}{Theorem}

\newtheorem{definition}{Definition}

\newcommand{\li}[1]{}
\newcommand{\linew}[1]{}
\newcommand{\linewnew}[1]{}

\newcommand{\kang}[1]{}
\newcommand{\ed}{}
\newcommand{\ednew}{}
\usepackage{authblk}

\pagebreak
\title{Runtime-Safety-Guided Policy Repair}
\author[1]{Weichao Zhou\thanks{zwc662@bu.edu}}
\author[2]{Ruihan Gao\thanks{GAOR0007@e.ntu.edu.sg}}
\author[3]{BaekGyu Kim\thanks{baekgyu.kim@toyota.com}}
\author[4]{Eunsuk Kang\thanks{eskang@cmu.edu}}
\author[1]{Wenchao Li\thanks{wenchao@bu.edu}}
\affil[1]{Boston University, Boston MA, USA}
\affil[2]{Nanyang Technological University, Singapore}
\affil[3]{Toyota Motor North America R\&D, Mountain View, CA, USA}
\affil[4]{Carnegie Mellon University, Pittsburgh, PA, USA}

\begin{document}

\maketitle              

\begin{abstract}
 We study the problem of policy repair for learning-based control policies in  safety-critical settings. 
 We consider an architecture where a high-performance learning-based control policy (e.g. one trained as a neural network) is paired with a model-based safety controller. The safety controller is endowed with the abilities to predict whether the trained policy will lead the system to an unsafe state, and take over control when necessary. 
 While this architecture can provide added safety assurances, intermittent and frequent switching between the trained policy and the safety controller can result in undesirable behaviors and reduced performance. 
 We propose to reduce or even eliminate control switching by `repairing' the trained policy based on runtime data produced by the safety controller in a way that deviates minimally from the original policy. 
 The key idea behind our approach is the formulation of a trajectory optimization problem that allows the joint reasoning of policy update and safety constraints. 
 Experimental results demonstrate that our approach is effective even when the system model in the safety controller is unknown and only approximated. 
\end{abstract}
\section{Introduction}
Data-driven methods such as imitation learning have been successful in learning control policies for complex control tasks~\cite{argall2009survey,codevilla2018end}. 
A major shortcoming that impedes their widespread usage in the field is that the learnt policies typically do not come with any safety guarantee.
It has been observed that when encountering states not seen in training, the learnt policy can produce unsafe behaviors
\cite{DBLP:journals/corr/AmodeiOSCSM16,ross2010efficient}. 

A common approach to mitigate the safety problem at runtime is to pair the learning-based controller\footnote{We use the terms `controller' and `control policy' (or simply `policy') interchangeably in this paper. The latter is more common in the machine learning literature.} (LC) with a high-assurance safety controller (SC) that can take over control in safety-critical situations, such as the Simplex architecture first proposed in \cite{simplex}.
The safety controller is tasked with predicting an impending safety violation and taking over control when it deems necessary.
Such controllers are often designed based on conservative models\ed\kang{slightly repetitive, since the phrase "designed conservatively" is already used in the previous sentence}, has inferior performance compared to its learning-based counterpart, and may require significant computation resources if implemented online (e.g. model predictive control). 
Moreover, frequent and intermittent switching between the controllers can result in undesirable behaviors and further performance loss. 

In this paper, we propose to \textit{leverage the runtime interventions carried out by the safety controller to repair the learnt policy}. 
We do not assume access to the original training data of the LC but we assume that the policy is parameterized, differentiable and given as a white-box. 
This means that while fine-tuning the LC from scratch is not possible, it is still possible to improve the controller based on new data that is gathered during deployment.\ed\kang{Maybe this is explained later, but why is (not) having access to original training data important?}
In particular, we introduce the concept of \textit{policy repair} which uses the outputs of the safety controller to synthesize new training data to fine-tune the LC for improved safety. 
Furthermore, we formalize a notion of \textit{minimal deviation} with respect to the original policy in order to mitigate the issue of performance degradation during policy repair. \ed\kang{Incomplete sentence?}
The main idea in \textit{minimally deviating policy repair} is the formulation of a trajectory optimization problem that allows us to \textit{simultaneously reason about policy optimization and safety constraints}. 
A key novelty of this approach is the synthesis of new safe `demonstrations' that are the most likely to be produced by the original unsafe learnt policy. 
In short, we make the following contributions.

\begin{itemize}
    \item[$-$] We formalize the problems of \textit{policy repair} and \textit{minimally deviating policy repair} for improving the safety of learnt control policies.
    \item[$-$] We develop a novel algorithm to solve the policy repair problem by iteratively synthesizing \ed\li{maybe we can further qualify the type of training data}new training data from interventions by the safety controller to fine-tune the learnt policy.
    \item[$-$] We demonstrate the effectiveness of our approach on case studies including a simulated driving scenario where the true dynamics of the system is unknown and is only approximated.
\end{itemize}



\section{Related Work}

\textbf{Model-based control} is a well-studied technique for controlling dynamical systems based on the modelling of the system dynamics. Algorithms such as iterative Linear Quadratic Regulator (iLQR)~\cite{6386025} have achieved good performance even in complex robotic control tasks. One important advantage of model-based control is its ability to cope with constraints on the dynamics, controls and states. Constrained Model Predictive Control \cite{maciejowski2002predictive} has been studied extensively and proven to be successful in solving collision avoidance problems~\cite{bareiss2013reciprocal,borrelli2004collision} as well as meeting complex high-level specifications~\cite{decastro2013guaranteeing}. 
In this paper, we utilize model-based control techniques to verify the existence of safe control as well as synthesize new training data to guide the policy learning\ednew\linew{a bit too high-level here; can we be more specific, e.g. MPSC and trajectory optimization?}. 


\textbf{Imitation learning} provides a way of transferring skills for a complex task from a (human) expert to a learning agent~\cite{ROB-053}. 
It has been shown that data-driven methods such as behavior cloning are effective\ednew\linew{effective?} in handling robotics and autonomous driving tasks~\cite{Pomerleau1988ALVINNAA,ross2011reduction} when an expert policy is accessible at training time.
Model-based control\ednew\linew{just model-based or model-based control? Ans: the second one} techniques have already been introduced to imitation learning to guide the policy learning process \cite{finn2016guided,6630832,inproceedings,Pereira2018MPCInspiredNN}. \add{Our work shares similarity with \cite{DBLP:journals/corr/abs-1802-05803} in using a model predictive controller to generate training examples. What distinguishes our work from theirs is that in \cite{DBLP:journals/corr/abs-1802-05803} the model predictive controller operates based on a given cost function whereas in our work we do not assume we know any cost function.}
An outstanding challenge in the imitation learning area is the lack of safety assurance during both training and final deployment. \add{Efforts on addressing this challenge include \cite{Zhang2017QueryEfficientIL,8968287}, where multiple machine learning models cooperate to achieve performance and safety goals. 
However, the learned models cannot not provide guarantees on runtime safety by themselves.
In fact, }even when the dynamical model is given, existing imitation learning algorithms \ed\kang{"as designed" or "by design"?}lack the means to incorporate explicit safety requirements. 
In this paper, we use imitation learning to formulate the problem of minimally deviating policy repair such that a repaired policy can match the performance of the original learnt policy while being safe.

\textbf{Safe Learning} research has experienced rapid growth in recent years. Many approaches consider safety requirement as constraints in the learning process. For example, \cite{achiam2017constrained,chow2018lyapunov} encodes safety as auxiliary costs under the framework of Constrained Markov Decision Processes (CMDPs). However, the constraints can only be enforced approximately.
\cite{chow2018lyapunov}\ednew\linew{name of the technique?} developed a Lyapunov-based approach to learn safe control policies in CMDPs but is not applicable to parameterized policy and continuous control actions. 
Formal methods have also been applied to certain learning algorithms for establishing formal safety guarantees.
In \cite{zhou2018safety}, safety is explicitly defined in probabilistic computational tree logic and a probabilistic model checker is used to check whether any intermediately learned policy meets the specification.
If the specification is violated, then a counterexample in the form of a set of traces is used to guide the learning process.
Providing assurance for runtime safety of learning-based controller has also garnered attention recently. \cite{fulton2018safe} combines offline verification of system models with runtime validation of system executions.  
In \cite{alshiekh2018safe}, a so-called shield is synthesized to filter out unsafe outputs from a reinforcement learning (RL) agent. It also promotes safe actions by modifying the rewards. 
A similar idea can be seen in \cite{Phan2019NeuralSA} where a so-called neural simplex architecture 
is proposed and an online training scheme is used to improve the safety of RL agents by rewarding safe actions. However,
in the context of RL, choosing the right reward is in general a difficult task, since incorrect choices often lead to sub-optimal or even incorrect solutions. 
In \cite{Cheng2019EndtoEndSR}, a model predictive approach is proposed to solve for minimum perturbation to bend the outputs of an RL policy towards asymptotic safety enforced by a predefined control barrier certificate. 
A similar idea also appears in \cite{Wabersich2018LinearMP} where robust model predictive control is used to minimally perturb the trajectories of a learning-based controller towards an iteratively expanding safe target set\ednew\linew{this sentence doesn't make sense to me}.
Our method differs from \cite{Cheng2019EndtoEndSR,Wabersich2018LinearMP} as we improve the runtime safety of the learning-based control while preserving its performance from an imitation learning perspective. 


\section{Preliminaries}
\label{prelim}

In this paper we consider a discrete-time control system $(X, U, f, d_0)$ where $X$ is the set of states of the system and $U$ is the set of control actions. The function $f:X\times U \rightarrow X$ is the dynamical model describing how the state evolves when an control action is applied, and $d_0:X\rightarrow \mathbb R$ is the distribution of the initial states\ednew\linew{multiple initial states? requiring sum to 1?}. 
By applying control actions sequentially, a trajectory, or a trace, $\tau=\{(x_t , u_t)|t=0, 1,\ldots\}$ can be obtained where $x_t, u_t$ are the state and control action at time $t$. 
In typical optimal control problems, a cost function $c:X\times U\rightarrow \mathbb R$ is explicitly defined to specify the cost of performing control action $u\in U$ in state $x\in X$. The cumulative cost along a trajectory $\tau$ can be calculated as $\underset{(x_t , u_t )\in\tau}{\sum}c(x_t , u_t )$. 
An optimal control strategy is thus one that minimizes the cumulative cost. 

\textbf{Model Predictive Control (MPC)} leverages a predictive model of the system to find a sequence of optimal control actions in a receding horizon fashion. It solves the optimal sequence of control actions for $T$ steps as in (\ref{1}) but only applies the first control action and propagates one step forward to the next state. Then it solves for a new sequence of optimal control actions in the next state. 

\resizebox{.91\linewidth}{!}{
  \begin{minipage}{\linewidth}
\begin{eqnarray}
&&\arg\underset{x_{0:T}, u_{0:T}}{\min} \sum^{T}_{t=0}c(x_t , u_t )\label{1}\\
s.t. && x_{t+1} = f(x_t , u_t )\qquad t=0, 1, 2,\ldots, T\label{2}\\
\nonumber
\end{eqnarray}
\end{minipage}
}

When the dynamics $f$ in constraint (\ref{2}) is nonlinear, the iterative Linear Quadratic Regulator (iLQR) algorithm \cite{Li2004IterativeLQ} applies a local linearization of $f$ along an existing trajectory which is called the nominal trajectory\ednew\linew{`nominal' is not very clear}. It computes a feedback control law via LQR ~\cite{kwakernaak1972linear}, which induces a locally optimal perturbation upon the nominal trajectory to reduce the cumulative cost. Formally, given a nominal trajectory $\{(x_0, u_0), ..., (x_{T}, u_{T})\}$, perturbations can be added to each state and control action in this trajectory, i.e. $x_t \rightarrow x_t  + \delta x_t , u_t \rightarrow u_t  +\delta u_t $. \ed\kang{Is there an error here? "(3) to (3)"} The relationship between $\delta x_t ,\delta u_t$ and $\delta x_{t+1}$ is locally determined by the dynamics as well as \add{the state and control actions in the nominal trajectory} as in (\ref{7}) where $\nabla_x f(x_t, u_t), \nabla_u f(x_t, u_t)$ are the partial derivatives of $f(x_t, u_t)$ w.r.t $x, u$. Meanwhile, based on the nominal trajectory, $\sum^{T}_{t=0}c(x_t, u_t)$ in the objective (\ref{1}) is substituted by $\sum^{T}_{t=0}c(\delta x_t  + x_t,\delta u_t + u_t )-c(x_t, u_t)$ \add{while the decision variables become $\delta x_{0:T}, \delta u_{0:T}$}. When adopting an online trajectory optimization strategy \cite{6386025}, the optimal control law has a closed form solution $\delta u_t = k_t + K_t\delta x_t$ in which $k_t, K_t$ are determined by the dynamics and the cumulative cost along the nominal trajectory. 

\resizebox{.91\linewidth}{!}{
  \begin{minipage}{\linewidth}
\begin{eqnarray}
x_{t+1} &=& f(x_t , u_t )\qquad x_{t+1} + \delta x_{t+1} = f(x_t +\delta x_t , u_t  + \delta u_t)\label{4}\label{5}\\
\delta x_{t+1}^T &\approx& \delta x_t ^T \nabla_x f(x_t , u_t )  +\delta u_t ^T\nabla_u f(x_t , u_t )\label{7}\\
\nonumber
\end{eqnarray}
\end{minipage}
}

A \textbf{control policy} in general is a function $\pi: X\rightarrow U$ that specifies the behavior of a controller in each state. Given a deterministic policy $\pi$, its trajectory can be obtained by sequentially applying control actions according to the outputs of $\pi$. 
Specifically, for an LC such as a deep neural network, the policy is usually parameterized and can be written as $\pi_{\theta}$ where the parameter $\theta$ belongs to some parameter set $\Theta$ (e.g. weights of a neural network). We assume that $\pi_\theta(x)$ is differentiable both in $x$ and $\theta$ \ed\kang{typo?}.

\textbf{Imitation learning} assumes that an expert policy $\pi_E$ (e.g. a human expert) can demonstrate on how to finish a desired task with high performance. The learning objective for an agent is to find a policy $\pi$ that matches the performance of $\pi_E$ in the same task\ednew\linew{not performance matching? this would be strict behavioral cloning, right?}. 
Traditional approaches such as behavioral cloning consider the 0-1 error $e(x_t,\pi_E;\pi)=\mathcal{I}\{\pi(x)\neq\pi_E(x)\}$ where $\mathcal{I}$ is an indicator function. \ed\kang{State here what $\mathcal{I}$ is} In this setting, an optimally imitating policy minimizes $\mathbb{E}_{x\sim d_{\pi_E}} [e(x, \pi_E; \pi)]$ where $d_{\pi_E}$ is state visitation distribution of $\pi_E$. From another perspective, the difference between $\pi$ and $\pi_E$ can be estimated based on their trajectory distributions. When the trajectory distribution $Prob(\tau|\pi_E)$ is known, one can empirically estimate and minimize the KL divergence $D_{KL}[\pi_E||\pi]$ by regarding $Prob(\tau|\pi)$ as the probability of $\pi$ generating trajectory $\tau$ under an additional Gaussian noise, i.e. $u_t\sim \mathcal{N}(\pi(x_t), \Sigma),\forall (x_t, u_t)\in\tau$. On the other hand, one can estimate and minimize the KL divergence $D_{KL}[\pi||\pi_E]$ by treating $Prob(\tau|\pi)$ as being induced from a Dirac delta distribution $u_t\sim \delta(\pi(x_t))\ \forall (x_t, u_t)\in\tau$. 
\ednew\linew{see commented out part in source; not clear what `Given' means}
Both KL-divergences are related to negative log-likelihoods.
\section{Runtime Safety Assurance}
\label{rtsa}

In this section we discuss the runtime safety issues of LCs and introduce our basic strategy for safe control. We consider a runtime safety requirement $\Phi$ for finite horizon $T$, such as `if the current state is safe at step $t$, do not reach any unsafe state within the next $T$ steps'. 
Temporal logic can be used to formally capture this type of safety requirements~\cite{maler2004monitoring,pnueli1977temporal}.  
Given an LC with a deterministic policy $\pi_{\theta}$ , if $\pi_\theta$ satisfies $\Phi$ globally, that is, at each time step along all its trajectories\ednew\linew{all the time is colloquial; be formal}, we denote it as $\pi_\theta\models \Phi$; otherwise $\pi_\theta\not\models\Phi$. 

We assume that for any satisfiable $\Phi$, there exists an SC, which we represent as $\pi^{safe}$, \ed\kang{A minor comment, but I found it slightly confusing to see $\pi$ (reserved for a policy) being used to refer to a controller} that checks at runtime whether $\Phi$ is satisfiable if\ed\li{filter is perhaps not the right verb here since it also produces safe controls} the output $\hat{u}=\pi_\theta(x)$ of the LC is directly applied. That is, whether there exists a sequence of control actions in the next $T-1$ steps such that $\Phi$ is not violated. If true, then the final output $\pi^{safe}(x, \pi_\theta(x))=\hat{u}$. Otherwise it overrides the LC's output with $\pi^{safe}(x, \pi_\theta(x))\neq \hat{u}$. We formally define the SC below.

\begin{definition}\label{def1}
Given a safety requirement $\Phi$, the corresponding SC is a mapping $\pi^{safe}$ from $X\times U$ to $U$. In each state $x\in X$, $\pi^{safe}(x, \pi_\theta(x))=\pi_\theta(x)$ iff $\Phi$ is satisfiable after applying the control action $\pi_\theta(x)$; otherwise, $\pi^{safe}$ intervenes by providing a substitute $\pi^{safe}(x,\pi_\theta(x))\neq \pi_\theta(x)$ to satisfy $\Phi$. 
\end{definition}

We use $\langle\pi_\theta, \pi^{safe}\rangle$ to represent the LC and SC pair. Obviously the trajectories generated by this pair satisfy $\Phi$ everywhere if $\pi^{safe}$ exists. 
There are multiple options
of implementing the SC
such as having a backup human safety driver or using automated reasoning. Depending on the safety requirement and task environment, the difficulty of implementing safe control varies. \ed\li{need to elaborate more here}  
In this paper, 
we assume that a dynamical model of $f$ is given, possibly constructed conservatively, and adopt a scheme known as Model Predictive Safe Control as detailed below.

\subsection{Model Predictive Safe Control}\label{rtsa_mpsc}

This scheme exploits the dynamical model to predict safety in the future. Depending on the safety requirement $\Phi$ considered, a function $\varphi:X\rightarrow \mathbb{R}$ can be defined to quantify how safe any state $x$ is, i.e. if $\varphi(x)\leq 0$, then $x$ is safe; otherwise $x$ is unsafe. Without loss of generality, we let the current step be $t=0$. Then the safety requirement can be translated into the constraints $\forall t\in \{1, 2, \ldots, T\}, \varphi(x_t)\leq 0$. 
After the LC provides a candidate control output ${u}_0=\pi_{\theta}(x_0)$, the SC first verifies the satisfiability of (\ref{12}) by using an MPC-like formulation as $(\ref{9})\sim(\ref{11})$. 

\resizebox{.91\linewidth}{!}{
  \begin{minipage}{\linewidth}
\begin{eqnarray}
&&\underset{x_{0:T}, u_{0:T}}{\text{min}}\qquad 0\label{9}\\
s.t.&&x_{t+1}=f(x_t, u_t)\qquad t=0, 1,2,\ldots, T-1\ \ \qquad\qquad\qquad\label{10}\\
&& \varphi(x_t)\leq 0\qquad t=1,2,\ldots, T\label{12}\\
&& u_0 =\pi_{\theta}(x_0)\label{11}\\
\nonumber
\end{eqnarray}
\end{minipage}
}

The formula differs from MPC in that it solves a feasibility problem to check the existence of a sequence of control actions satisfying the constraints.\ed\kang{Maybe this is a standard formulation of feasibility in MPC, but I am having a hard time parsing (5). Here the goal is to check the existence of a sequence of controller actions for the next T time steps s.t. the property is satisfied, correct? I guess the actions are implicitly quantified as part of (5)?  ignore my comment if this is standard} It is easier to solve than optimal control since optimality is not required here. \ed\li{can we say something along the lines of the feasibility problem being easier to solve or the optimization objective being hard to formulate correctly?}If this problem is feasible, that is, $(\ref{10})\sim(\ref{11})$ can be satisfied at the same time. Then $\pi_{\theta}(x_0)$ is deemed safe and the final output is $\pi^{safe}(x_0, \pi_\theta(x_0))=\pi_{\theta}(x_0)$. Otherwise, the SC solves another feasibility problem which is the same as $(\ref{9})\sim (\ref{12})$ and has (\ref{11}) removed because the unsafe candidate control action $\pi_{\theta}(x_0)$ is to be substituted.
Note that it is possible that (\ref{12}) is unsatisfiable, in which case there is no feasible solution. This means a safety violation is inevitable based on the given model, but the SC can predict such outcome $T$ steps in advance and more drastic actions (e.g. physically changing the model) may be applied to prevent an accident from occurring.\ed\li{how to terminate?} 
If a feasible solution to $(\ref{9})\sim (\ref{12})$ can be obtained, we let $\pi^{safe}(x_0, \pi_\theta(x_0))=u_0$ and use this solved $u_0$ to evolve the system to the next state. \ed\kang{I am slightly confused by this; is this "first control output" the result of the second feasibility problem (with (8) removed)?}

\ed\li{I would first introduce MPSC and then discuss the connection with STL-based MPC.}
\li{No need to explicitly say 'in formal methods'. Use a paragraph header in bold to highlight the connection to STL-based MPC.}
There have been works on model predictive control of cyber-physical systems subject to formal specifications in signal temporal logic (STL) and its probabilistic variant~\cite{raman2014model,sadigh2016safe}. Techniques have been proposed to synthesize safety constraints from formal specifications to accommodate optimal control of continuous systems and to reason about safety under uncertainty. In the semantics of STL, $\varphi$ can be viewed as the negation of the robustness satisfaction value.\ed\kang{If $\varphi$ represents the robustness, I found it slightly strange that $\varphi \leq 0$ represents the property being satisfied  (usually positive robustness values represent satisfaction, and negative values a violation of the property). It's probably not a big deal, but someone familiar with robustness in STL may wonder the same}

In this paper, at the beginning of each time step, before solving the feasibility problem  $(\ref{9})\sim (\ref{11})$, we forward simulate the policy $\pi_{\theta}$ for $T$ steps. If the simulated trajectory satisfies the safety constraint (\ref{12}) already, then there is no need to query the SC at all. Otherwise, we use the constrained iLQR approach from \cite{8317745} to solve the feasibility problem. This approach treats the simulated trajectory as nominal trajectory and iteratively update the nominal trajectory. Also, this approach turns the safety constraint $(\ref{12})$ into a penalty $\sum^{T}_{t=0} exp(M_t \psi(x_t))$ with sufficiently large $\{M_t\}^{T}_{t=0}$. And the penalty is added to the objective.  By using this approach, even if the feasibility problem cannot be solved, at least a low-penalty solution can be provided.



\add{\noindent \textbf{Monitoring overhead.}} \ed\li{this sentence needs revision}Model Predictive Safe Control (MPSC) can provide assurance for a variety of runtime safety requirements. However, it can be more expensive to implement in practice compared to an LC due to the need to repeatedly solve a (nonlinear) optimization online as opposed to performing inference on a neural network\li{use proper latex for citation}  [Wuet al.2019]. \ed\kang{I don't know I agree with this statement as currently phrased. I'd expect the SC to have lower complexity than the ML-based controller and thus is actually easier/cheaper to build/validate. Maybe what you want to emphasize is the performance cost introduced by the SC}  Frequently using an SC to both verify safety and solve safe control at runtime can be computationally taxing for the entire control system. \add{For instance, say the LC's inference time is $t_{LC}$, the time for solving $(\ref{9})\sim (\ref{11})$ is $t^{(1)}_{SC}$ and the time for solving $(\ref{9})\sim (\ref{12})$ is $t^{(2)}_{SC}$. At each step, forward simulation of the LC for $T$ steps takes at least $T*t_{LC}$ time. 
If \eqref{12} is violated in the forward simulation, the SC
would be invoked and the total overhead will grow to $T*t_{LC} + t^{(1)}_{SC}$. 
If the problem based on LC's candidate control output is infeasible and the SC is required to intervene with a substitute control value, then 
the SC will have to solve another MPC-like problem and
the overhead will grow to 
$T*t_{LC} + t^{(1)}_{SC}+t^{(2)}_{SC}$.} \ed\li{needs revision} 
Thus, it would be more economical to have an inherently safe LC such that the SC is less triggered. Motivated by this, we propose to \textit{repair} the LC so that it becomes safer and requires less intervention from the SC. In the next section, we formally introduce the policy repair problem and provide a solution to it.

\section{Policy Repair}

\begin{figure}[!htb]
\centering
\includegraphics[scale=.30]{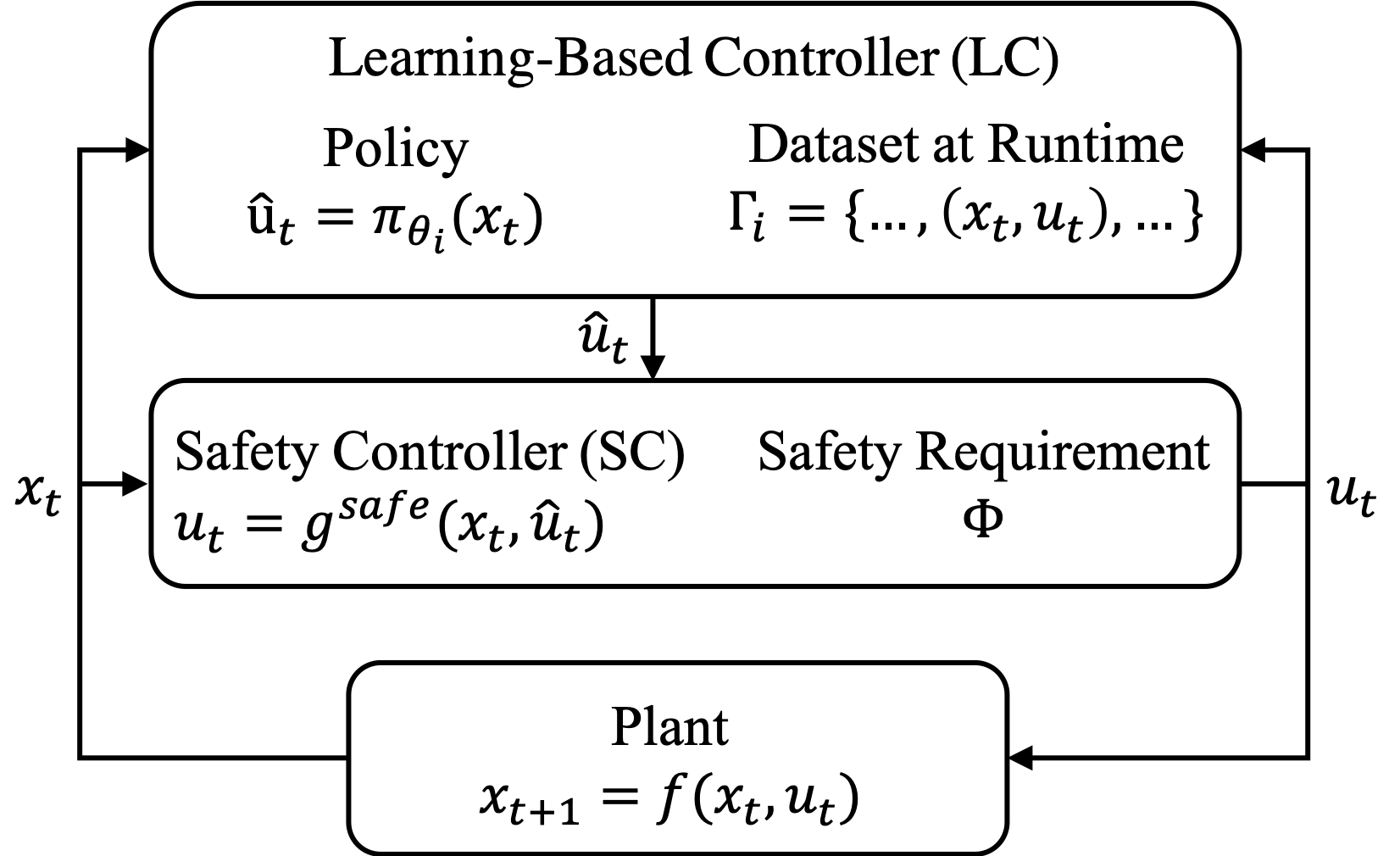}
\caption{Architecture of pairing LC's policy $\pi_{\theta}$ with an SC $\pi^{safe}$.}\label{fig:p1}
\vspace{-3mm}
\end{figure}

\ed\li{we should have a subsection title on the synthesis of 'new' safe demonstrations}
\noindent
We first give a formal definition of the policy repair problem below.

\begin{definition}\label{def2}
\ed\li{needs revision}
Given a deterministic policy $\pi_\theta$ paired with an SC $\pi^{safe}$ as defined in Definition~\ref{def1},
\textbf{policy repair} is the problem of finding a new policy $\pi_{\theta^*}$ such that  ${\theta^*}=\underset{\theta\in\Theta}{\arg\min}\ \mathbb{E}_{x\in X}[\mathcal{I}\{\pi^{safe}(x, \pi_{\theta}(x)) = \pi_{\theta}(x)\}]$ where $\mathcal{I}\{\cdot\}\in\{0, 1\}$ is an indicator function.
\end{definition}

Definition 2 implies that a repaired policy generates safe controls most of the time and thus the SC rarely intervenes. 
\li{needs better transition to next sentence: why is it intuitive?}
The first idea is to treat controls generated by the SC as repairs at specific states, and then use this data to repair the whole policy.
A solution based on this idea is described as follows. 

\subsection{Naive Policy Repair\ed\li{runtime retraining seems to suggest that you are doing the retraining at runtime} \ed\kang{Same comment as Wenchao's: It wasn't clear whether the training was happening offline or online (I assumed offline); this could be made more explicit earlier in the paper (e.g., intro)}}

\ed\li{introduce the idea of retraining/fine-tuning first}
During the execution of the LC and SC pair $\langle\pi_\theta, \pi^{safe}\rangle$, due to the presence of the SC, all the generated traces are safe. The basic idea of the naive policy repair approach is to let the unsafe LC learn from the interventions produced by the SC. Specifically, we iteratively execute the LC and SC pair to generate new safe traces. After each iteration, the state-action pairs in all the previously generated traces are used as training data to update the policy of the LC. We present the steps in Algorithm 1 and illustrate them with a high-level diagram in Fig.~\ref{fig:p1}, where $\Gamma_i$ is the set of traces of the $\langle\pi_{\theta_i}, \pi^{safe}\rangle$ pair at the $i^\text{th}$ iteration. We use supervised learning to fine-tune the policy parameter to minimize the expected error $\mathbb{E}_{(x,u)\sim \cup\Gamma_i} [e(x, u;\pi_{\theta})]$ as in line $9$ of Algorithm 1. Note that at this stage, with a slight abuse of notation, we view $\Gamma_i$ as a data set containing $(x,u)$ pairs.
In line $5\sim 7$, if the SC no longer intervenes, then we have a high confidence that the current policy is safe. 
According to the law of large numbers, 
this confidence increases with increasing number of sampled traces.
The algorithm also terminates if a maximum iteration number is reached, in which case the SC may still intervene and the policy repair is only partially successful. 

 
\vspace{-3mm}
\begin{algorithm}
\caption{Naive\_Policy\_Repair}
\begin{algorithmic}[1]
\STATE \textbf{Input} an initial policy $\pi_{\theta_0}$; \\
\STATE \textbf{Given} an SC $\pi^{safe}$; iteration parameter $N>0$; policy parameter set $\Theta$. \\
\FOR{iteration $i=0$ to $N$}
\STATE Run the $\langle\pi_{\theta_i}, \pi^{safe}\rangle$ pair to generate a set $\Gamma_i$ of trajectories. 
\IF {$\forall(x, u)\in \Gamma_i, u=\pi_{\theta_i}(x)$}
\STATE $\pi^{safe}$ never intervenes $\Rightarrow\pi_{\theta_i}\models\Phi$ with high probability. 
\STATE \textbf{return $\pi_{\theta_i}, \Gamma_i$}
\ENDIF
\STATE {${\theta_{i+1}}=\underset{\theta\in\Theta}{\arg\min}\ \mathbb{E}_{(x,u)\sim \cup^i_{j=0} \Gamma_j} [e(x, u;\pi_{\theta})]$}
\ENDFOR
\RETURN $\pi_{\theta_N}, \emptyset$
\end{algorithmic}
\end{algorithm}
\vspace{-1mm}

\subsection{Analysis of Performance Degradation due to SC Intervention}\label{ana} \ed\linewnew{need a better title}
In this section, we analyze the performance degradation due to the application of safe controls from the SC and use it to motivate the study of better policy repair strategies.
We assume that the initial learnt policy $\pi_{\theta_0}$ is given as a white-box and its parameter $\theta_0$ has already been optimized for the control task. \kang{Related to my earlier comment, why do you assume this?} Inspired from lemma 1 in \ednew\linew{can't we mentioned DAgger explicitly?}\cite{schulman2015trust}, we analyze the performance degradation of naive policy repair in a fixed-horizon task with maximum step length $H$. Recall the definition of cost function $c$ in Section \ref{prelim}\ednew\linew{use section reference}. Without loss of generality, we simplify it into a function of state, that is, from $c(x,u)$ to $c(x)$ and normalize it to the range $[0, 1]$. 
We use $\eta(\pi)=\mathbb{E}_{\tau\sim \pi}[\sum^H_{t=0}c(x_t)]$ to denote the expected cumulative cost of following a policy $\pi$ from initialization to step $H$. Define the value function $V_\pi(x_t)=\mathbb{E}_{x_t, u_t, x_{t+1}\ldots\sim\pi}[\sum^H_{l=t}c(x_l)]$ as the expected cost accumulated by following $\pi$ after reaching state $x_t$ at step $t$ till step $H$. Define the state-action value function $Q_\pi(x_t, u_t)=\mathbb{E}_{x_t, x_{t+1}, u_{t+1}\ldots\sim\pi, u_t}[\sum^H_{l=t}c(x_l)]$ as the expected cost accumulated by executing $u_t$ in state $x_t$, then following $\pi$ henceforth til step $H$. We use an advantage function\ednew\linew{advantage?}  $A_\pi(x_t, u_t) = Q_\pi(x_t, u_t) -V_\pi(x_t)$ to evaluate the additional cost incurred by applying control action $u_t$ in $x_t$ instead of adhering to $\pi$. Based on the lemma 1 \ednew\linew{refer to Proof number}in \cite{schulman2015trust} for infinite-horizon scenario, we have the equation (\ref{trpo}) for any two policies $\pi, \hat{\pi}$ in finite-horizon scenario. 

\resizebox{.91\linewidth}{!}{
  \begin{minipage}{\linewidth}
\begin{eqnarray}
\mathbb{E}_{\tau\sim \hat{\pi}}[\sum^H_{t=0} A_\pi(x_t, u_t)]&=&\mathbb{E}_{\tau\sim\hat{\pi}}[\sum^H_{t=0} c(x_t) + V_\pi(x_{t+1}) - V_\pi(x_t)]\nonumber\\
= \mathbb{E}_{\tau\sim\hat{\pi}}[-V_\pi(x_0) + \sum^H_{t=0} c(x_t)]&=& \mathbb{E}_{x_0\sim d_0}[-V_\pi(x_0)] + \mathbb{E}_{\tau\sim\hat{\pi}}[\sum^H_{t=0} c(x_t)]=\eta(\hat{\pi}) - \eta(\pi)\qquad\label{trpo}
\end{eqnarray}
\end{minipage}
}

Assuming that $\eta(\pi_{\theta_0})$ is the minimum for the desired task, i.e. $\pi_{\theta_0}$ is the optimal policy with respect to a cost function $c$\li{added}, we bound the additional cost $\eta(\pi^{safe})-\eta(\pi)$ incurred by possible interventions of $\pi^{safe}$.

\begin{theorem}
Given a $\langle\pi_{\theta_0},\pi_{safe}\rangle$ pair, let $\epsilon_{1}, \epsilon_2$ and $\epsilon_3$ be the probability of $\langle\pi_{\theta_0},\pi_{safe}\rangle$ generating a $H$-length trajectory where $\pi^{safe}(x, \pi_{\theta_0}(x))\neq \pi_{\theta_0}(x)$ happens in at least one, two and three states respectively. 
Then, $\eta(\pi^{safe}) - \eta(\pi_{\theta_0})  \leq \epsilon_1 H  + \epsilon_2 (H-1)+ \frac{\epsilon_3(H-1)H}{2}$. (Proof in Appendix)
\end{theorem}
\li{I moved $\eta(\pi_{\theta_0})$ to the left hand side.}
\kang{I am not sure I entirely understand what this theorem says (after reading a couple of times); perhaps an informal explanation/intuition would help} 

\begin{proof}
Define $e(x)\in\{0, 1\}$ as the probability of the safety controller intervening in state $x$. Let $p_{<t}$ represent the probability of the safety controller never intervening before step $t$. Then we use $d_t(x_t, u_t)$ to represent the probability of generating $x_t, u_t$ at step $t$ conditioned on $p_{<t}$, while using $d'_t(x_t, u_t)$ to represent the probability  of generating $x_t, u_t$ at step $t$ but conditioned on $1-p_{<t}$. Let $p_{>t}$ be the probability of the safety controller never intervening after step $t$ conditioned on the fact that the safety controller intervenes not only at step $t$ and also for at least one time before step $t$. Then obviously $\epsilon_1 = \sum^H_{t=0} p_{<t}\mathbb{E}_{(x_t, u_t)\sim d_t} [e(x_t)]$, $\epsilon_2 = \sum^{H}_{t=1} (1 - p_{<t})\mathbb{E}_{(x_t, u_t)\sim d'_t} [e(x_t)]p_{>t}$ and $\epsilon_3\geq (1 - p_{<t})\mathbb{E}_{(x_t, u_t)\sim d'_t} [e(x_t)](1-p_{>t})\ \forall t\in\{0, 1, 2, \ldots, H\}$. Note that $A_{\pi_{\theta_0}}(x_t, u_t) = 0$ in states where $u_t = \pi^{safe}(x_t, \pi_{\theta_0}(x_t))=\pi_{\theta_0}(x_t)$ while  $A_{\pi_{\theta_0}}(x_t, u_t) \geq 0$ in states where $u_t = \pi^{safe}(x_t, \pi_{\theta_0}(x_t))\neq\pi_{\theta_0}(x_t)$ due to the optimality of $\pi_{\theta_0}$ under the current cost function $c$. In addition, $A_{\pi_{\theta_0}}(x_t, u_t)\leq (H-t)(\underset{x\in X}{\max}\ c(x) - \underset{x\in X}{\min}\ c(x))=H-t$ for all $x_t\in X, t\in\{0, 1, \ldots, H\}$. Then we use those facts and assumptions to derive the theorem as below based on (\ref{trpo}).\end{proof}

\resizebox{.91\linewidth}{!}{
\begin{minipage}{\linewidth}
\begin{eqnarray}
&&\eta(\pi^{safe})\nonumber\\
&=&\eta(\pi_{\theta_0}) + \mathbb{E}_{\tau\sim\pi^{safe}}[\sum^H_{t=0}A_{\pi_{\theta_0}}(x_t, u_t)]\nonumber\\
&=& \eta(\pi_{\theta_0}) + \sum^H_{t=0}\mathbb{E}_{x_t, u_t\sim\pi^{safe}}[e(x_t)A_{\pi_{\theta_0}}(x_t, u_t)]\nonumber\\
&=& \eta(\pi_{\theta_0}) + \sum^H_{t=0}p_{<t}\mathbb{E}_{x_t, u_t\sim d_t} [ e(x_t)A_{\pi_{\theta_0}}(x_t, u_t)]  + \sum^H_{t=1} (1 - p_{<t})\mathbb{E}_{x_t, u_t\sim d'_t}[e(x_t) A_{\pi_{\theta_0}}(x_t, u_t)]p_{>t} \nonumber\\
&& +  \sum^{H-1}_{t=1}(1 - p_{<t})\mathbb{E}_{x_t, u_t\sim d'_t}[e(x_t)A_{\pi_{\theta_0}}(x_t, u_t)](1 - p_{>t}) \nonumber\\
&\leq& \eta(\pi_{\theta_0}) + \sum^H_{t=0}p_{<t}\mathbb{E}_{x_t, u_t\sim d_t} [ e(x_t)(H-t)]  + \sum^H_{t=1} (1 - p_{<t})\mathbb{E}_{x_t, u_t\sim d'_t}[e(x_t) (H-t)p_{>t}]\nonumber\\
&&+  \sum^{H-1}_{t=1} (H-t)\epsilon_3 \nonumber\\
&\leq& \eta(\pi_{\theta_0}) + H \sum^H_{t=0}p_{<t}\mathbb{E}_{x_t, u_t\sim d_t} [ e(x_t)]  + (H-1)\sum^H_{t=1} (1 - p_{<t})\mathbb{E}_{x_t, u_t\sim d'_t}[e(x_t)p_{>t}]+  \sum^{H-1}_{t=1} (H-t)\epsilon_3 \nonumber\\
&\leq& \eta(\pi_{\theta_0}) + \epsilon_1 H  + \epsilon_2 (H-1)+ \frac{\epsilon_3(H-1)H}{2}
\end{eqnarray}
\end{minipage}
}

The theorem shows the additional cost can grow quadratically in $H$ when the probability of multiple interventions from the SC becomes higher. 
The implication of this is that even if the repaired policy $\pi_{\theta^*}$ replicates $\pi^{safe}$ with zero error, 
the repaired policy can still suffer from significant performance degradation.
Since the training error is non-zero in practice, $\pi_{\theta^*}(x)\neq \pi_{\theta_0}(x)$ may happen in more states where $\pi^{safe}(x,\pi_{\theta_0}(x))\neq \pi_{\theta_0}(x)$. 
One major challenge in mitigating this performance loss is that the training information of $\pi_{\theta_0}$, especially the cost function $c$, could be unknown. 
In the next section, we describe our approach of repairing a policy so that it also minimally deviates from the original one. 

\subsection{Minimally Deviating Policy Repair via Trajectory Synthesis}\label{mdpr}

We firstly formally define the minimally deviating policy repair problem.\li{the following definition is a definition of a policy being minimally deviating but not a definition of the problem}
\begin{definition}
Given an initial policy $\pi_{\theta_0}$ and an SC $\pi^{safe}$ as defined in Definition 1, \textbf{minimally deviating policy repair} is the problem of 
finding a policy $\pi_{\theta^*}$ where  $\theta^*=\underset{\theta\in\Theta}{\arg\min} \mathbb{E}_{x\sim d_{\pi_\theta}} [e(x, \pi_{\theta_0}; \pi_\theta)]$ subject to $\pi^{safe}(x, \pi_{\theta}(x))=\pi_{\theta}(x), \forall x\in X$.
\end{definition}

\li{Two questions here. (1) $\pi_\theta$ needs to be a distribution but you also require equality in the constraint. (2) Is there a way to define this equivalently but more intuitively without using KL divergence? This is especially important since in the following section, you use a special treatment for the deterministic policy $\pi_\theta$.}


Informally, the objective of this repair problem is to reduce the chance of $\pi_{\theta^*}(x) \neq \pi_{\theta_0}(x)$ while maintaining the safety of $\pi_{\theta^*}$. 
Observe that the error term $e(\cdot)$ in Definition 3 resembles the one in an imitation learning setting. Then minimizing the expected error can be viewed as imitating $\pi_{\theta_0}$.  On the other hand, the equality constraint in Definition 3 can be understood as requiring $\pi_{\theta^*}$ to satisfy (\ref{12}) at all steps in all its trajectories. Hence, the minimally deviating policy repair is essentially a problem of optimizing an imitation learning objective with safety constraints. The major challenge is that, the decision variable for the imitation learning objective is the policy parameter $\theta$ while for safety constraints (\ref{12}) it is the state $x$. 

\begin{figure}[!htb]
\vspace{-3mm}
\begin{minipage}{.24\linewidth}
\centering
\subfigure{\label{p4}\includegraphics[scale=.3]{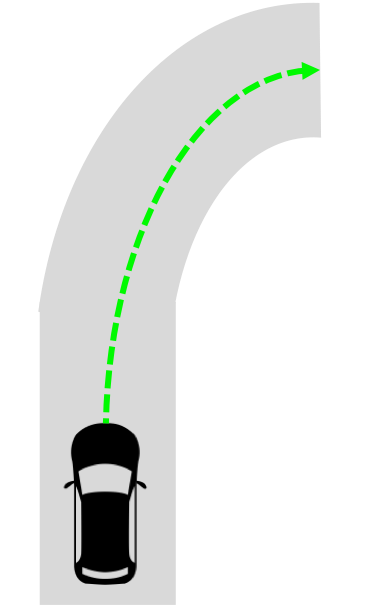}}\\
(a)
\end{minipage}
\begin{minipage}{.24\linewidth}
\centering
\subfigure{\label{p5}\includegraphics[scale=.3]{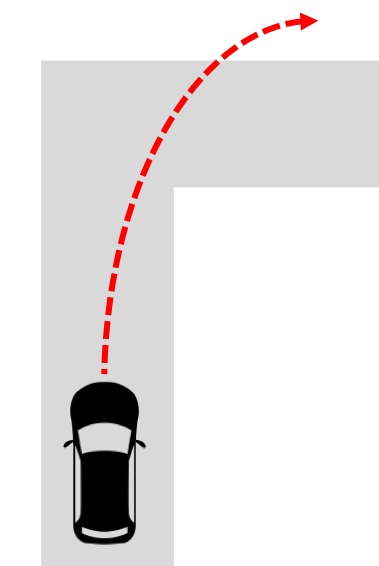}}\\
(b)
\end{minipage}
\begin{minipage}{.24\linewidth}
\centering
\subfigure{\label{p6}\includegraphics[scale=.3]{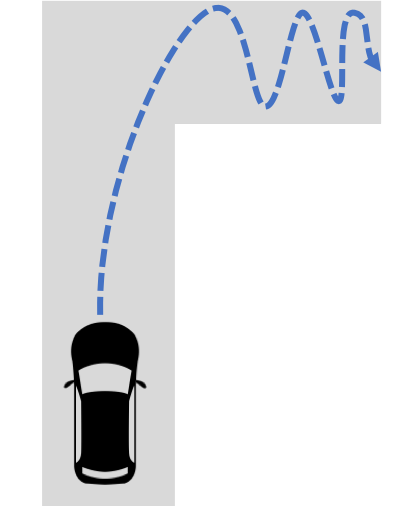}}\\
(c)
\end{minipage}
\begin{minipage}{.24\linewidth}
\centering
\subfigure{\label{p7}\includegraphics[scale=.3]{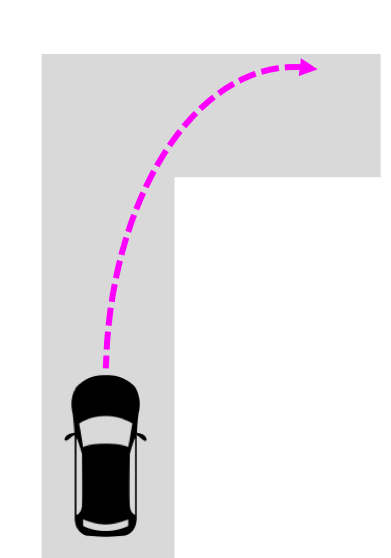}}\\
(d)
\end{minipage}
\caption{
\add{ (a) The grey area is the lane. The green dashed curve is the trajectory of the vehicle. (b) The red dashed curve is the trajectory of the initial policy. (c) The blue dashed curve is the trajectory of the policy and safety controller pair. (d) The magenta dashed curve is the trajectory produced by the repair policy that deviates minimally from the original one.
}}
\vspace{-5mm}
\end{figure}

\add{
We use a simple example below to illustrate our problem setting and desired solution. Consider a policy that was trained to steer a vehicle around a specific corner as shown in Fig.~\ref{p4}. 
When deployed in a slightly different environment as shown in Fig.~\ref{p5}, the policy fails to keep the vehicle inside the lane.
Fig.~\ref{p6} illustrates that with the basic simplex setup as shown in Fig.~\ref{fig:p1}, although the safety controller manages to keep the vehicle inside the lane, frequent switching between the two controllers can lead to undesirable behaviors such as an oscillating trajectory.}
\add{Fig.~\ref{p7} shows a more desirable trajectory produced by a new policy trained using minimally deviating policy repair.
\li{I am not sure whether you want to show trajectories that are used to train the policy or trajectories that are produced by the final trained policy.}
Our approach to the problem stated in Definition 3 is to `imitate' the original policy by first synthesizing and then learning from new trajectories that are similar to ones produced by the original policy but instead do not violate the safety requirements. 
The synthesis algorithm works by iteratively improving the trajectories produced by a naively repaired policy such as the one in Fig.~\ref{p6} until trajectories such as the one in Fig.~\ref{p7} are obtained. The improvement is achieved by solving a trajectory optimization problem \highlight{of which the objective} is transformed from the imitation learning objective in Definition 3. We mainly focus on showing such transformation in the rest of this section. 
}

As mentioned in Section~\ref{prelim}, to solve an imitation learning problem, we can minimize the KL-divergence which is related to maximal log-likelihood, i.e. $\underset{\theta\in\Theta}{\arg\min}\ D_{KL}[\pi_\theta||\pi_{\theta_{0}}]= \underset{\theta\in\Theta}{\arg\max}\ \mathbb{E}_{\tau\sim\pi_\theta} [\log Prob(\tau|\pi_{\theta_{0}})]$. Note that $Prob(\tau|\pi_\theta)$ is induced from a Dirac Delta distribution $u\sim\delta(\pi(x))$ and $Prob(\tau|\pi_{\theta_{0}})$ is carried out by adding to $\pi_{\theta_{0}}$ an isotropic Gaussian noise $\mathcal{N}(0,\Sigma)$ with diagonal $\Sigma=\sigma^2 I$. When a finite set $\Gamma$ of trajectories of $\pi_\theta$ is obtained, the log-likelihood is equivalent to (\ref{28}). \kang{This is where I started to get a little lost, possibly due to my lack of familiarity with imitation learning. Before you jump into heavy math, can you give a high-level overview of how this policy repair technique works? Perhaps start by describing Algorithm 2 first and then go into the math?}

\resizebox{.91\linewidth}{!}{
  \begin{minipage}{\linewidth}
\begin{eqnarray}
    &&\mathbb{E}_{\tau\sim\pi_\theta} [\log Prob(\tau|\pi_{\theta_{0}})]\approx \frac{1}{|\Gamma|}\sum_{\tau\in\Gamma} \log Prob(\tau|\pi_{\theta_{0}})\nonumber\\
    \propto&&\sum_{\tau\in \Gamma} \log \{\prod_{(x_t, u_t)\in \tau}exp[-\frac{(\pi_\theta(x_t) - \pi_{\theta_0}(x_t))^T \Sigma^{-1}(\pi(x_t,\theta)- \pi_{\theta_0}(x_t))}{2}]\}\nonumber\\
    \propto&& -\frac{1}{2}\sum_{\tau\in \Gamma} \sum_{(x_t, u_t)\in\tau}||\pi_\theta(x_t)-\pi_{\theta_0}(x_t)||^2_2\label{28}
\end{eqnarray}
\end{minipage}
}

Suppose that at iteration $i\geq 1$, a safe policy $\pi_{\theta_{i}}$ is obtained and executed to generate a set $\Gamma_i$ of safe traces. Define $l_{x_t,\pi_{\theta_i}}=\frac{1}{2}||\pi_{\theta_0}(x_t)-\pi_{\theta_i}(x_t)||^2_2$ and $J_{\Gamma_i}(\pi_{\theta_i})=\sum_{\tau\in \Gamma_i} \sum_{(x_t, u_t)\in\tau}l_{x_t,\pi_{\theta_i}}$. To decrease $J_{\Gamma_i}$, a new policy parameter $\theta_{i+1}=\theta_i + \delta\theta_i$ can be obtained by solving $\delta\theta_i=\underset{\delta\theta}{\arg\min}\ J_{\Gamma_i}(\pi_{\theta_i + \delta\theta}) - J_{\Gamma_i}(\pi_{\theta})$. We further use the Gauss-Newton step \cite{nocedal2006numerical} to expand this as shown in (\ref{27}) below.

\resizebox{.91\linewidth}{!}{
  \begin{minipage}{\linewidth}
\begin{eqnarray}
&&\underset{\delta\theta}{\arg\min}\ \delta\theta^T \nabla_\theta J_{\Gamma_i}(\pi_{\theta_i}) + \frac{1}{2} \delta\theta^T \nabla_\theta J_{\Gamma_i}(\pi_{\theta_i})\nabla_\theta J_{\Gamma_i}(\pi_{\theta_i})^T\delta\theta\nonumber\\
=&&\underset{\delta\theta}{\arg\min}\ \sum_{\tau\in \Gamma_i}\sum_{(x_t, u_t)\in \tau } \delta\theta_i\nabla_\theta \pi_{\theta_i}(x_t)\nabla_{\pi_{\theta_i}} l_{x_t,\pi_{\theta_i}}\nonumber\\
&&\qquad\qquad\qquad +\ \frac{1}{2}  \delta\theta_i^T\nabla_\theta \pi_{\theta_i}(x_t)\nabla_{\pi_{\theta_i}} l_{x_t,\pi_{\theta_i}} \nabla_{\pi_{\theta_i}} l_{x_t,\pi_{\theta_i}}^T\nabla_\theta\pi_{\theta_i}(x_t)^T\delta\theta_i\qquad \label{27}
\end{eqnarray}
\end{minipage}
}

 We note that the changes of the policy control output $u_t=\pi_{\theta_i}(x_t)$ at arbitrary state $x_t$ can be locally linearized as from (\ref{16}) to (\ref{17}). 

\resizebox{.91\linewidth}{!}{
  \begin{minipage}{\linewidth}
\begin{eqnarray}
&&u_t + \delta u_t = \pi_{\theta_i + \delta \theta_i}(x_t + \delta x_t)\qquad u_t=\pi_{\theta_i}(x_t)\label{16}\\
&&\delta u_t^T - \delta x_t^T \nabla_x \pi_{\theta_i}(x_t) \approx  \delta \theta_i^T \nabla_\theta \pi_{\theta_i}(x_t)\label{17}
\end{eqnarray}
\end{minipage}
}
\vspace{2mm}

It implies that due to $\delta\theta_i$, each trajectory $\tau=\{(x_0, u_0), (x_1, u_1), \ldots\}$ of $\pi_{\theta_i}$ is approximately perturbed by ${\delta\tau}=\{(\delta x_0, \delta u_0), (\delta x_1, \delta u_1), \ldots\}$. Motivated by the fact that $\pi_{\theta_i+\delta\theta_i}$ is safe if all of the trajectories are still safe after such perturbations, we optimize w.r.t the trajectory perturbations $\delta\tau$'s instead of $\delta\theta_i$ by exploiting the relation between each $(\delta x_t, \delta u_t)\in \delta\tau$ and $\delta\theta_i$ as in (\ref{17})\ed\linewnew{$\delta\tau$ doesn't show up in \ref{17}, so it is not clear what ``their" are referring to.}. Interpolating the RHS of (\ref{17}) in (\ref{27}), we obtain a trajectory optimization problem (\ref{29}) with linear and quadratic costs as shown in $(\ref{_LP})\sim(\ref{QP_})$. Note that this trajectory optimization problem treats the trajectories from $\Gamma_i$ as nominal trajectories and solves for optimal perturbations to update those nominal trajectories. 
Local linearization is used to derive the dynamics constraints as in (\ref{dynamics}) for each noiminal trajectory. 
By adding the safety constraints (\ref{safety}), the trajectories can remain safe after adding the solved perturbations. 
Here, we use the constrained iLQR approach from \cite{8317745} to resolve this constrained trajectory optimization problem. 

\vspace{-2mm}
\resizebox{.91\linewidth}{!}{
  \begin{minipage}{\linewidth}
  \begin{eqnarray}
\underset{\{\delta x_{0:H}, \delta u_{0:H}\}}{\arg\min}&\ &\frac{1}{4|\Gamma_i|}\sum_{\tau\in \Gamma_i} \sum_{(x_t, u_t)\in\tau}
\begin{bmatrix} 1\\\delta x_t \\ \delta u_t\end{bmatrix}^T 
\begin{bmatrix} 
0 & Q_x^T &  Q_u^T\\ 
Q_x & Q_{xx} & Q_{xu} \\
Q_u & Q_{xu}^T & Q_{uu}
\end{bmatrix} 
\begin{bmatrix}1\\\delta x_t \\ \delta u_t\end{bmatrix} \label{QP}\label{29}\qquad\quad\ \ \ \\
\nonumber\\
where\ Q_x\ &=& -2\nabla_x \pi_{\theta_i}(x_t)\nabla_{\pi_{\theta_i}} l_{x_t,\pi_{\theta_i}} \label{_LP}\\
Q_u\ &=& 2\nabla_{\pi_{\theta_i}} l_{x_t,\pi_{\theta_i}}\label{LP_}\\
Q_{xx}&=&\nabla_x \pi_{\theta_i}(x_t) \nabla_{\pi_{\theta_i}} l_{x_t,\pi_{\theta_i}}\nabla_{\pi_{\theta_i}} l_{x_t,\pi_{\theta_i}}^T \nabla_x \pi_{\theta_i}(x_t)^T \label{_QP}\\
Q_{xu}&=& \nabla_x \pi_{\theta_i}(x_t) \nabla_{\pi_{\theta_i}} l_{x_t,\pi_{\theta_i}}\nabla_{\pi_{\theta_i}}l_{x_t,\pi_{\theta_i}}^T\qquad\qquad\\
Q_{uu}&=& \nabla_{\pi_{\theta_i}} l_{x_t,\pi_{\theta_i}}\nabla_{\pi_{\theta_i}} l_{x_t,\pi_{\theta_i}}^T \label{QP_}\\
s.t.\qquad\delta x_{t+1}^T &=&  \delta x_t ^T \nabla_x f(x_t , u_t )  +\delta u_t ^T\nabla_u f(x_t , u_t )\qquad t=0, 1,\ldots, H-1\label{dynamics}\\
\varphi(x_t &+ &\delta x_t)\leq 0\qquad t=0,1,2,\ldots, H\label{safety}
\end{eqnarray}
\end{minipage}
}
\vspace{1mm}

One major benefit of this formulation is that \textit{imitation learning objective and safety constraints can be reasoned at the same time via optimal control}. As the optimization is now separable, $(\ref{QP})\sim(\ref{dynamics})$ provide a lower bound for (\ref{27}). By solving the linear equations (\ref{17}), $\delta\theta_i$ can be inferred from the solved perturbations $\{\delta x_{0:H}, \delta u_{0:H}\}$, and then be used to modify $\theta_i$. Alternatively, $\pi_{\theta_{i}+\delta\theta_i}$ can be obtained by training $\pi_{\theta_{i}}$ with the trajectories induced from $\{x_{1:H}+\delta{x_{1:H}}, u_{1:H}+\delta{u_{1:H}}\}$.

\begin{figure}[!htb]
\centering
\includegraphics[scale=.30]{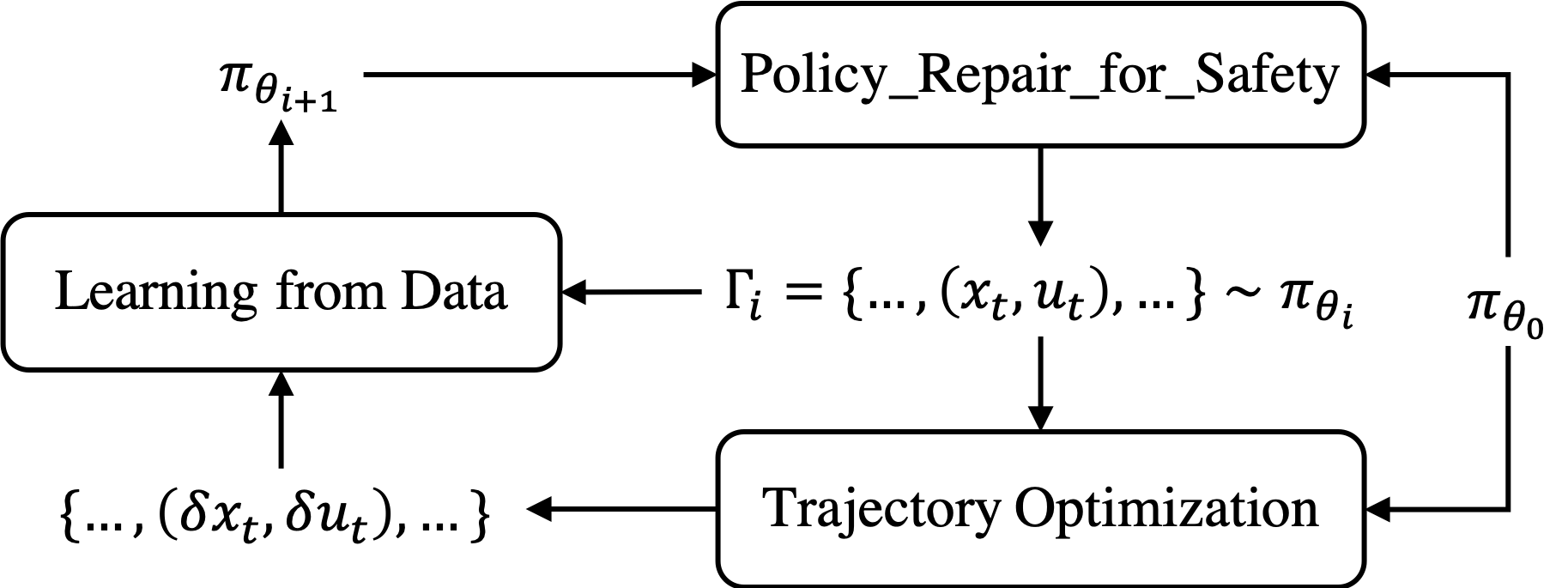}
\caption{\li{caption needs revision}Key steps in our minimally deviating policy repair algorithm. $\pi_{\theta_{0}}$ refers to the initial, learnt policy.\kang{I think this figure needs to be at the beginning of this section}}\label{fig:p2}
\end{figure}

The key steps of this iterative approach are shown in Fig.\ref{fig:p2} and the details are in Algorithm 2.\ed\kang{As with Algorithm 1, it would be helpful to walk the reader through how this algorithm works by referring to specific lines} As indicated by line 2 and 6, Algorithm 1 is used to find safe policies and generate safe nominal trajectories. This is because safe nominal trajectories guarantee that the trajectory optimization problem $(\ref{QP})\sim(\ref{safety})$ has feasible solutions, e.g. $\delta x=0, \delta u=0$. 
We terminate Algorithm 2 if Algorithm 1 fails to output a set of safe trajectories. In each iteration, we solve for the trajectory perturbations in line 4 and use them to update the policy as shown in line 5. The algorithm ends in line 7 if the trajectory optimization step no longer helps in decreasing the deviation.

\begin{algorithm}
\caption{Policy\_Repair\_for\_Minimal\_Deviation}
\begin{algorithmic}[1]
\STATE \textbf{Given} an initial learnt policy $\pi_{\theta_0}$; iteration parameters $\epsilon\in[0, 1], N>1$. \\
\STATE \textbf{Initialization} Obtain $\pi_{\theta_1}, \Gamma_1$ from $\text{Naive\_Policy\_Repair}(\pi_{\theta_0})$ via \textbf{Algorithm 1}; \\
\textbf{if} $\Gamma_{1}$ is $\emptyset$, \textbf{then return} \textit{fail} \\
\FOR{iteration $i=1$ to $i=N$}
\STATE Solve the optimal $\{\delta x_{0:H}, \delta u_{0:H}\}$ from $(\ref{QP})\sim(\ref{safety})$.
\STATE Solve $\delta \theta_i$ via (\ref{17}) and let $\theta_{i+1}= \theta_{i} + \delta{\theta_i}$. \\
Alternatively, search for ${\theta_{i+1}}=\underset{\theta\in\Theta}{\arg\min}\ \mathbb{E}_{(x,u)\sim\Gamma_i}[e(x+\delta x, u+\delta u;\pi_{\theta})]$ by training $\pi_{\theta_i}$ with $\{(x+\delta x, u+\delta u)|(x,u)\in\Gamma_i\}$.
\STATE Obtain $\pi_{\theta_{i+1}}, \Gamma_{i+1}$ from $\text{Naive\_Policy\_Repair}(\pi_{\theta_{i+1}})$ via \textbf{Algorithm 1}; \\
\textbf{if} $\Gamma_{i+1}$ is $\emptyset$, \textbf{then return} $\pi_{\theta_{i}}$\\
\STATE\textbf{if} {$|J_{\Gamma_{i+1}}(\pi_{\theta_{i+1}})-J_{\Gamma_{i}}(\pi_{\theta_{i}})|\leq \epsilon$}, \textbf{then return} $\pi_{\theta_{i+1}}$\\
\ENDFOR
\RETURN $\pi_{\theta_N}$
\end{algorithmic}
\end{algorithm}

\add{\textbf{Complexity analysis.} The major time complexity of Algorithm 2 will be accounted for by solving the quadratic programming (QP) in $(\ref{QP})\sim(\ref{safety})$. Since the cost \eqref{QP} is convex as indicated by \eqref{28}, if the constraint \eqref{safety} is also convex, the complexity of solving such QP can be polynomial \cite{doi:10.1137/1.9781611970791}; otherwise, it can be NP-hard \cite{Pardalos1991QuadraticPW}. The trajectory optimization in line $4$ needs to be solved only once off-line at the beginning of each iteration based on} {the safe}  \add{trajectories collected from previous iteration. In our experiments, the trajectory optimization is solved in a receding horizon manner as an MPC. In this case, the QP will be solved repeatedly over time to determine an appropriate sequence of control outputs. 
The nominal trajectories are obtained at each step by forward simulating the policy for a few steps ahead. 
The total time budget will be the same as the standard MPC. 
Besides the trajectory optimization, the time complexity of policy updating in line $5$ is either the same as that of solving an approximated linear equation \eqref{17} or training a neural network in a standard supervised manner.}

\section{Experiments}
\ed\li{need to state upfront what we will be evaluating} \kang{+1 to Wenchao's comment; what exactly are you trying to demonstrate with these experiments?}
We perform two case studies to evaluate the effectiveness of our proposed approach. 
The key metrics of evaluation are (1) safety of the repaired policy and (2) performance preservation with respect to the original policy.
\subsection{Mountaincar}

\begin{figure}[!htb]
\begin{minipage}{.5\linewidth}
\centering
\subfigure{\label{fig8}\includegraphics[scale=.45]{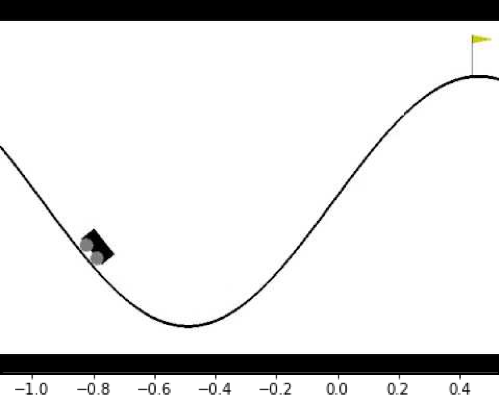}}\\
(a)
\end{minipage}%
\begin{minipage}{.5\linewidth}
\centering
\subfigure{\label{fig9}\includegraphics[scale=.3]{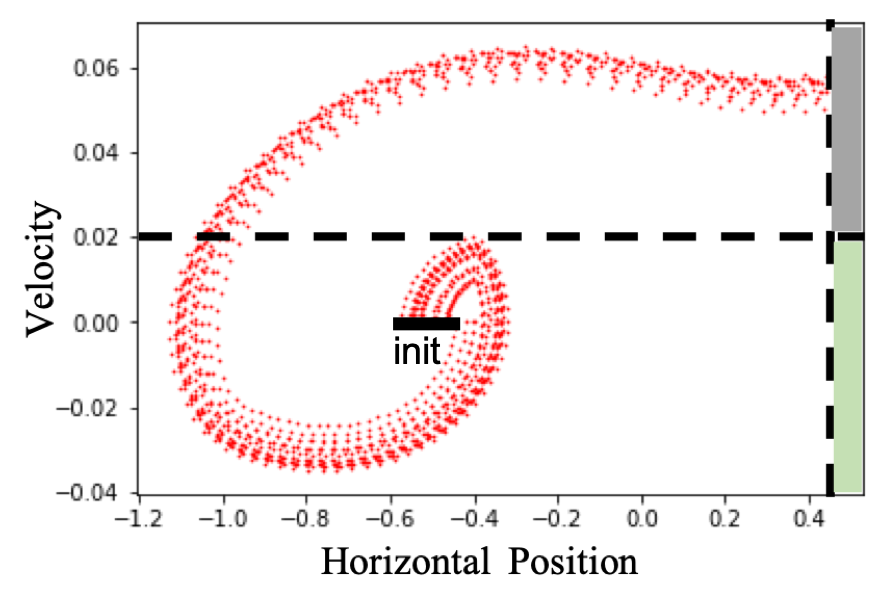}}\\
(b)
\end{minipage}
\begin{minipage}{.5\linewidth}
\centering
\subfigure{\label{fig10}\includegraphics[scale=.3]{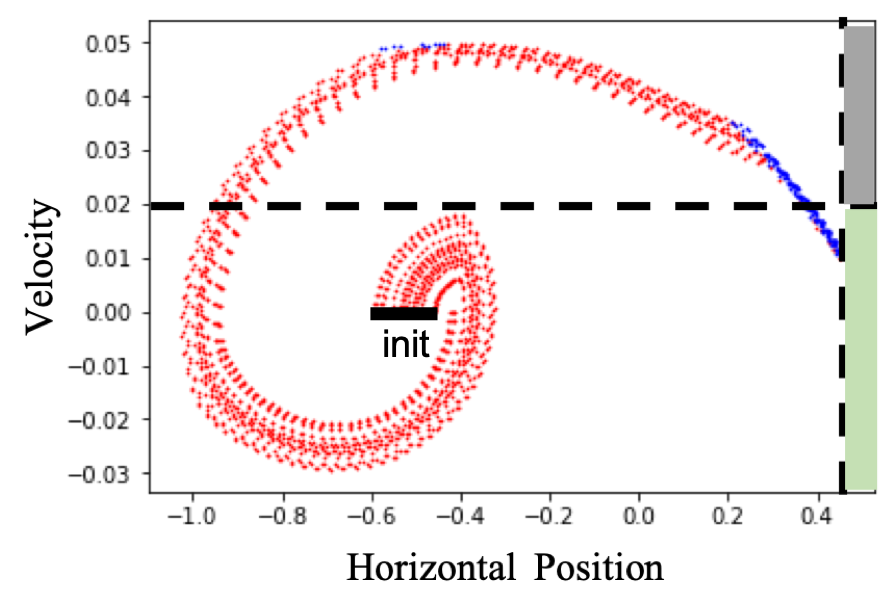}}\\
(c)
\end{minipage}%
\begin{minipage}{.5\linewidth}
\centering
\subfigure{\label{fig12}\includegraphics[scale=.3]{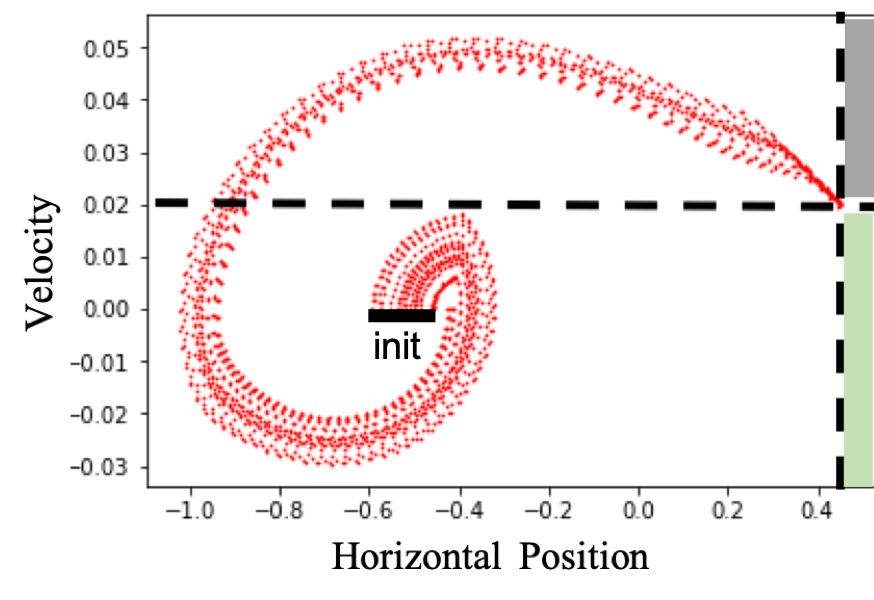}}\\
(d)
\end{minipage}
\caption{(a) The mountaincar environment. (b) The red patterns represent a set of trajectories produced by executing the initial policy. The y-axis indicates the velocity and the x-axis indicates the horizontal position of the car. The car reaches the right mountain top in $83.8$ steps on average with velocity higher than the safety threshold ($0.02$).\li{indicate initial position in the figure?} (c) The interventions by the SC are indicated by the blue dots. A naively repaired policy takes the $89.3$ steps on average to reach the mountaintop. (d) A minimally deviating repaired policy accomplishes the same task in $84.9$ steps on average without violating the safety requirement.\li{is there another way to do the comparison, since the percentage difference is small if we consider the whole trajectory?}}
\end{figure}

Our first case study is Mountaincar\footnote{https://gym.openai.com/envs/MountainCarContinuous-v0/}\ednew\linew{citation?}, as shown in Fig.\ref{fig8}. \li{removed Continuous but we should note elsewhere that the dynamics is continuous} In this environment, the goal is to push an under-powered car from the bottom of a valley to the mountain top on the right with as few steps as possible. 
The state ${\bf x}=[p, v]$ includes the horizontal position $p\in[-1.2, 0.6]$ and the velocity $v\in[-0.07, 0.07]$ of the car. The control $u\in[-1.0, 1.0]$ is the force to be applied to the car. The car has a discrete-time dynamics that can be found in the source code the simulator. For the LC, we train a neural network policy \ednew\linew{not clear who trains it?}via the Proximal Policy Optimization (PPO) algorithm \cite{DBLP:journals/corr/SchulmanWDRK17}. The neural network takes the state variables as input and generates a distribution over the action space. An additional layer is added at the end of the network to calculate the expected action. In Fig.$\ref{fig9}\sim (d)$, the x and y axes indicate the horizontal position and the velocity respectively. The car starts from a state randomly positioned 
within $[-0.6,-0.4]$ as indicated by the black line above `init'. The step length  \ednew\linew{averaged over how many runs?} for the PPO-trained policy to reach the mountain top ($p\geq 0.45$) is $83.8$ averaged over $1000$ runs. 

Now consider the safety requirement \textit{`velocity $v$ should not exceed $0.02$ when reaching the mountain top at $p \geq 0.45$'}\ed\li{changed $x$ to $p$}. 
The goal states and unsafe states are indicated by the green and grey areas in Fig.\ref{fig9}. 
It can be observed that the PPO-trained policy does not satisfy this requirement as all the red trajectories in Fig.\ref{fig9} end up at $p=0.45$ with $v>0.02$. 
Then an SC is implemented \ednew\linew{how is it implemented?}by following the Model Predictive Safe Control scheme introduced in Section \ref{rtsa_mpsc}. The function $\varphi(x)$ in (\ref{12}) evaluates whether the state $x$ is in the grey unsafe area. 
The LC and SC pair generates the red trajectories in Fig.\ref{fig10}. The blue dots indicate the intervention of the SC. While implementing Algorithm 1 and Algorithm 2, in each iteration we collect $20$ trajectories in the trajectory set $\Gamma_i$. Algorithm 1 produces a naively repaired policy that can reach the green area with $89.3$ steps on average. 
When using the minimally deviating policy repair algorithm (Algorithm 2), the resulting policy produces the red trajectories in Fig.\ref{fig12}. 
It shows that in all the runs the resulting policy satisfies the safety requirement and in addition the SC does not intervene. In terms of performance, the policy reaches the green area with only $84.9$ steps on average, which is much closer to the performance of the original policy.

\subsection{Traction-Loss Event in Simulated Urban Driving Environment}

\begin{figure}[!htb]
\begin{minipage}{.37\linewidth}
\centering
\subfigure{\label{fig16}\includegraphics[scale=.128]{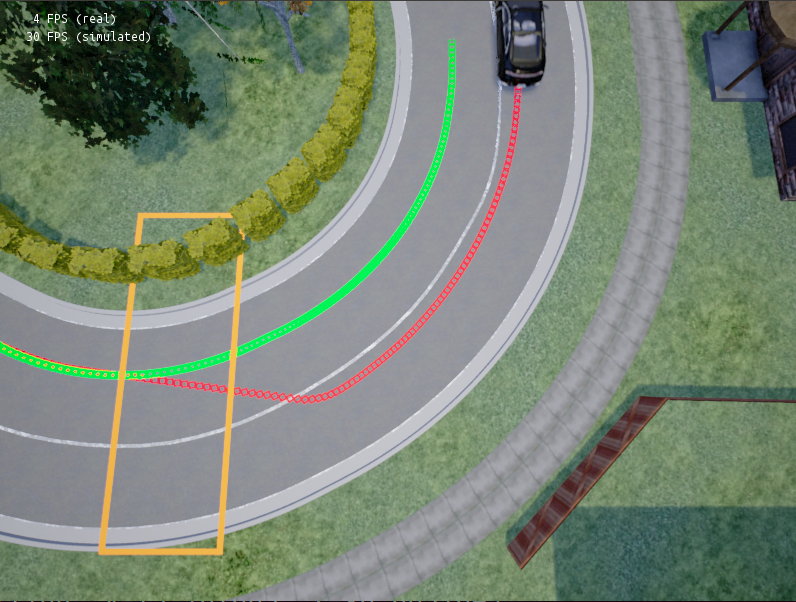}}\\
(a)
\end{minipage}%
\begin{minipage}{.3\linewidth}
\centering
\subfigure{\label{fig17}\includegraphics[scale=.128]{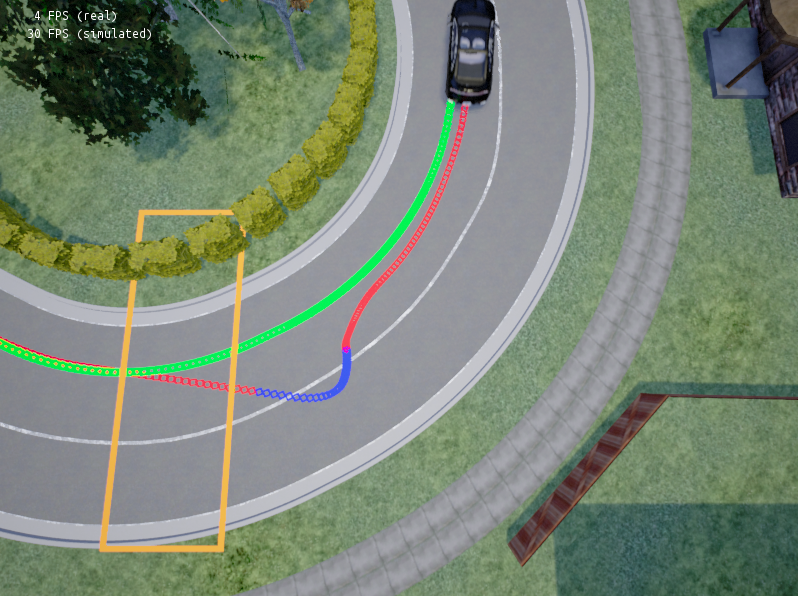}}\\
(b)
\end{minipage}%
\begin{minipage}{.3\linewidth}
\centering
\subfigure{\label{fig18}\includegraphics[scale=.15]{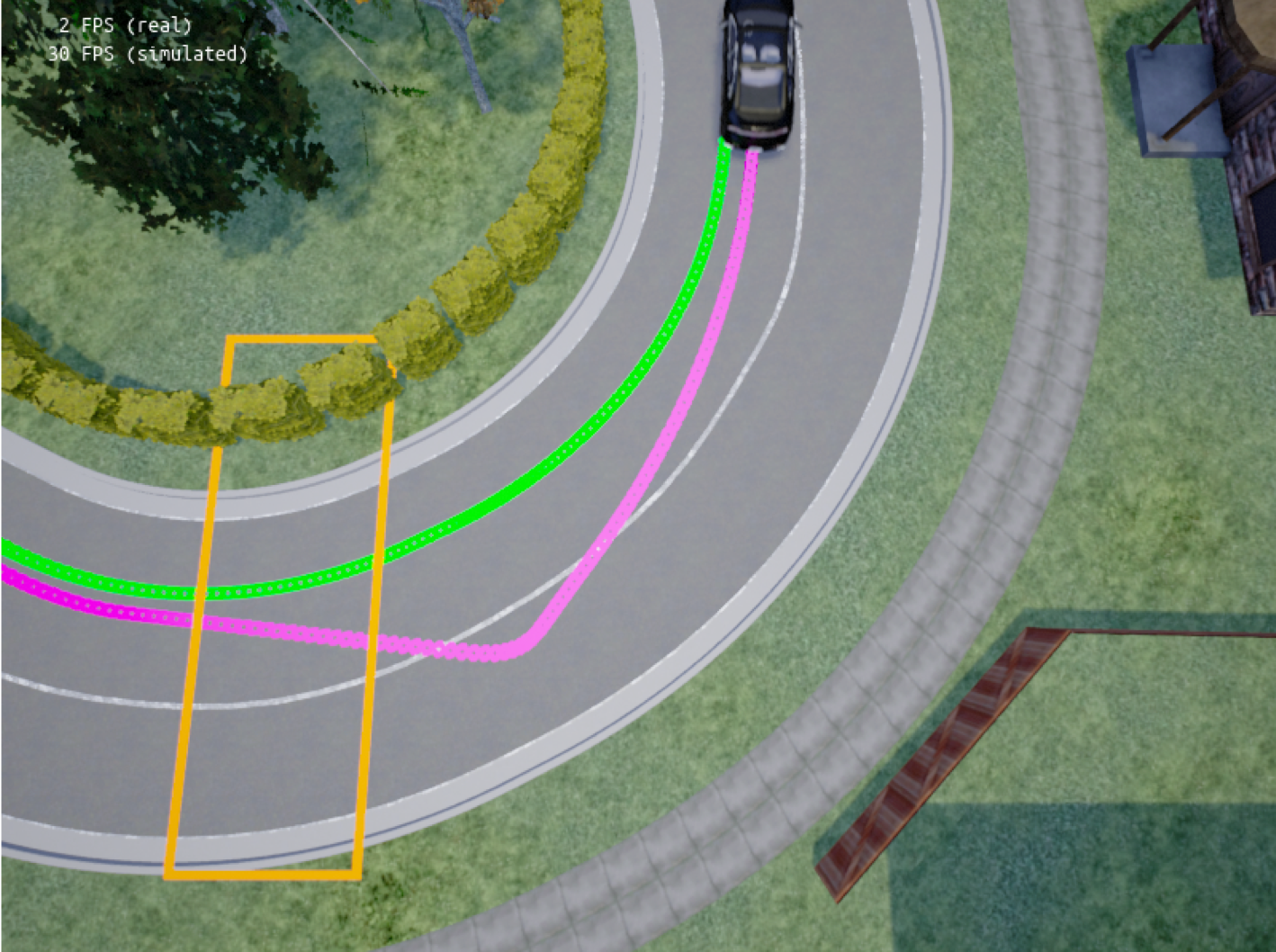}}\\
(c)
\end{minipage}
\begin{minipage}{.37\linewidth}
\centering
\subfigure{\label{fig18_1}\includegraphics[scale=.23]{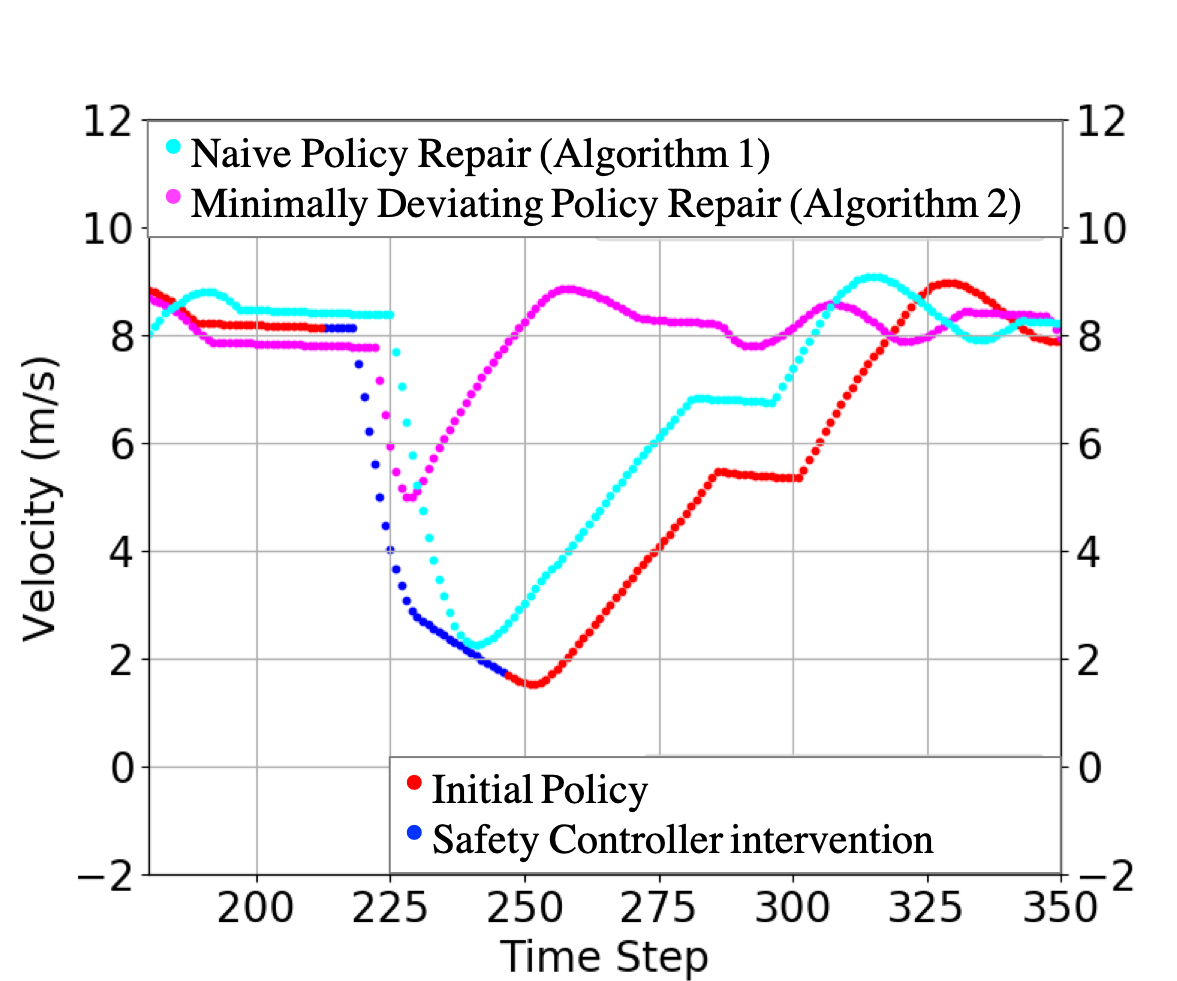}}\\
(d)
\end{minipage}%
\begin{minipage}{.3\linewidth}
\centering
\subfigure{\label{fig19}\includegraphics[scale=.15]{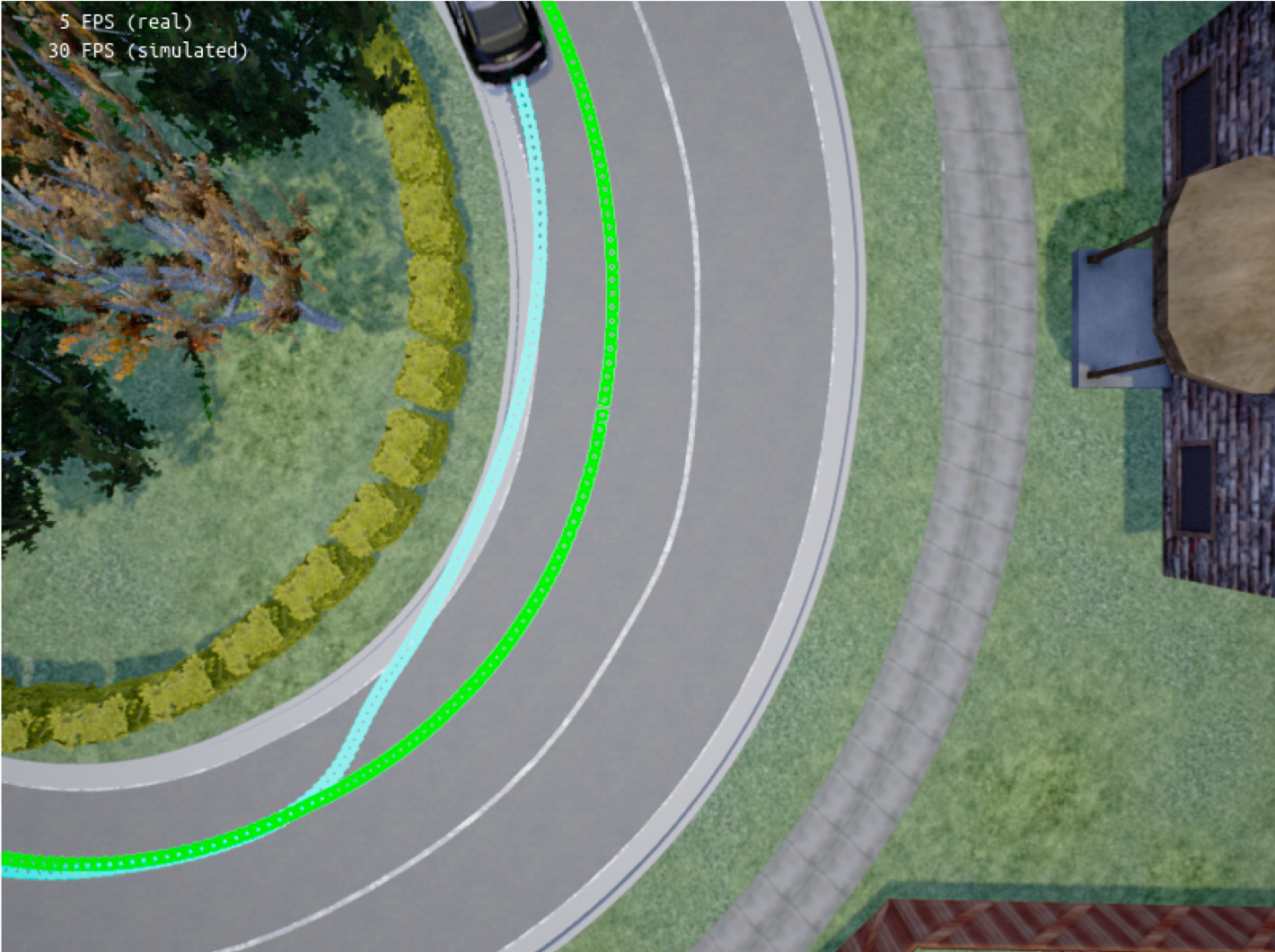}}\\
(e)
\end{minipage}%
\begin{minipage}{.3\linewidth}
\centering
\subfigure{\label{fig20}\includegraphics[scale=.15]{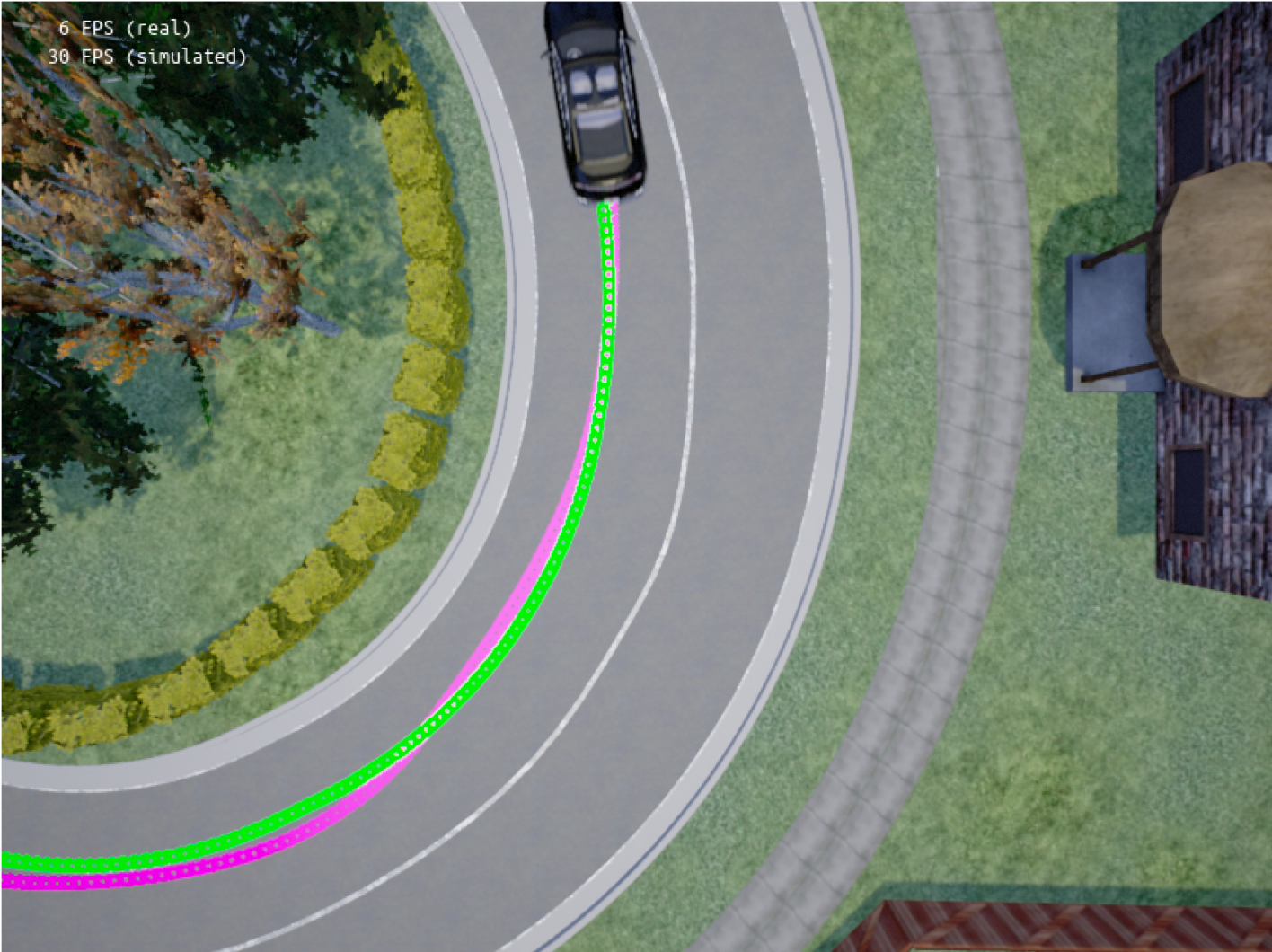}}\\
(f)
\end{minipage}
\caption{\add{The green trajectories represent normal trajectories of the car when there is no traction loss. The occurrence of the traction-loss event is indicated by the yellow rectangle. (a)Red trajectory: the initial policy fails to correct itself from skidding. (b) With interventions by the SC (the blue segment), the vehicle manages to return to the lane. (c) Magenta trajectory: policy repaired via Algorithm 2 corrects itself from skidding  and does so better than using the SC. (d) The Y-axis represents velocity of the car and the X-axis represents time steps. The red curve indicates that the initial policy is in control and the blue segments represents the interventions from the SC. The cyan curve is generated by a policy repaired via Algorithm 1. The magenta curve is generated by a minimally deviating policy repaired via Algorithm 2. (e) Cyan trajectory: after the traction-loss area is removed, the naively repaired policy drives the vehicle towards the center of the roundabout, crosses the inner lane boundary for a short time and then returns onto the lane. (f) Magenta trajectory: after the traction-loss area is removed, by using Algorithm 2, the vehicle adheres to the mid of the lane.}}
\vspace{-5mm}
\end{figure}

In this experiment, we show that our approach is effective even with an approximate dynamical model. The environment is in an open urban driving simulator, CARLA \cite{pmlr-v78-dosovitskiy17a}, with a single ego car on an empty road.
The state variables include position, velocity and yaw angle of the car and the control variables include acceleration and steering angles. 
We use a simple bicycle model from \cite{8317745} to approximate the unknown dynamical model of the car. \add{The model simulates a discrete-time system where the control actions are supplied to the system at an interval of $0.03s$.} 
For the LC, an initial neural network policy is trained with data collected from manually driving the car on different empty roads while maintaining a speed of $8m/s$ and keeping the car to the middle of the lane. During testing, we put the vehicle in a roundabout as shown in Fig.\ref{fig16} where the white curves are the lane boundary. The starting and finishing lines are fixed. The safety requirement can be described informally as \textit{`once the vehicle crosses outside a lane boundary, the controller should drive the vehicle back to the original lane within $5$ seconds'}. 




The initial, learnt policy drives the car well in the roundabout, as shown in  Fig.\ref{fig16}. 
We then consider an unforeseen traction-loss event, as shown by the yellow rectangle in Fig.\ref{fig16} where the friction is reduced to $0$ (e.g. an icy surface). 
As a result, the vehicle skids out of the outer lane boundary. The initial policy alone does not satisfy the safety requirement, as it keeps driving the vehicle outside the lane boundary after the traction-loss event, as shown by the red trajectory in Fig.\ref{fig16}. 
An SC is implemented by following the Model Predictive Safe Control scheme introduced in Section \ref{rtsa_mpsc}. The function $\varphi(x)$ in (\ref{12})  checks whether the distance between the vehicle and the middle of the lane is larger than half of the lane width. In Fig.\ref{fig17}, the blue segment indicates the interventions of the SC. 
It shows that due to the coupling of the LC and SC, the vehicle satisfies the safety requirement as it moves back to the lane.\add{When Algorithm 1 and 2 are executed, the parameter $\epsilon$ is set to $0.001$.} 
For every intermediate policy in each iteration, $10$ trajectories are collected in its trajectory set $\Gamma$.\add{It takes $5$ iterations for Algorithm 1 to synthesize a safe policy that does not require the SC to intervene. Starting with this safe policy, Algorithm 2 runs for $25$ iterations before termination.} The magenta trajectory in Fig.\ref{fig18} is from the minimally deviating policy repaired via Algorithm 2. Obviously the policy is able to correct itself without any intervention from the SC.\ednew\linew{significance?} In Fig.\ref{fig18_1}, we compare the velocities of the vehicles controlled by different policies.\ednew\linew{don't be repetitive if notations/colors are already explained in the caption. focus on discussing the results and pointing out the significance of the experiments. see the last paragraph as an example}
It can be observed that the velocities of all policies drop drastically due to traction-loss at around step $220$. The minimally deviating repaired policy performs the best in restoring the velocity back to $8m/s$. It is worth noting that velocity stability is important from the viewpoint of passenger comfort.\ednew\linew{perhaps we can relate this speed change to passenger comfort}

\begin{table}
\centering
\begin{minipage}{1.\linewidth}\centering
\begin{tabular}[t]{ p{3cm} || C{1.8cm} | C{1.8cm} | C{1.8cm} | C{1.8cm}}
 \hline
  &Avg. Speed($m/s$) & Lowest Speed($m/s$)&  Avg.\qquad\qquad Distance($m$) & Tot. Steps ($0.03s/step$) \\
 \hline
Initial Policy (No Traction-Loss Event)&8.0 &7.1 & 0.27 &396 \\
 \hline
Initial Policy& 8.0 & 5.2 & 1.7 & 420\\
Initial Policy w/ SC& 7.1&1.2 &  0.81 & 454 \\
Algorithm 1&7.5& 2.4 & 1.1 &  440 \\ 
\textbf{Algorithm 2}&\textbf{7.9}& \textbf{5.2} &\textbf{0.63} &  \textbf{413}\\
 \hline
\end{tabular}
\newline
\begin{tabular}[t]{ p{3cm} || C{1.8cm} | C{1.8cm} | C{1.8cm} | C{1.8cm}}
 \hline
  &Var. Speed & Std. Speed Change & Var. Distance & Std. Distance Change\\ 
 \hline
Initial Policy (No Traction-Loss Event)&0.53 & 0.074 & 0.10& 0.0096\\
 \hline
Initial Policy&0.79 & 0.16 & 4.4 & 0.026 \\
Initial Policy w/ SC&2.1 &0.17 & 1.0 &0.033\\ 
Algorithm 1& 2.4 & 0.17 &  1.4 & 0.042 \\
\textbf{Algorithm 2}&\textbf{0.73}& \textbf{0.15} &\textbf{1.0} & \textbf{0.033}\\
 \hline
\end{tabular}
\end{minipage}
\vspace{2mm}
\caption{\linewnew{can we compute the loss for init policy with SC as well? Need to clarify whether $J_\Gamma$ is computed with respect to the init policy without traction loss or with traction loss.}Avg. Speed: average speed of the vehicle in each run. Lowest Speed: the lowest speed since the vehicle firstly reaches $8m/s$ in each run. Aveg. Distance: the average distance between the vehicle and the middle of the lane at each step. Tot. Steps: the total number of steps that the vehicle outputs control actions in one run. 
\add{Var. Speed: the variance of the speed at each step in each run. Std. Speed Change: the standard deviation of the speed changes between consecutive steps. Var. Distance: the variance of the distance between the vehicle and the middle of the lane at each step. Std. Distance Change: the standard deviation of the distance (from vehicle to the middle of the lane) changes between consecutive steps.} Initial policy is tested before and after the traction-loss area is placed. The initial policy and SC pair is tested after the traction-loss event occurs. `Algorithm 1' and `Algorithm 2' respectively refer to the policies repaired via those two algorithms.}\label{table1}
\end{table}

We summarize the results in Table.\ref{table1}. \add{The performances of the algorithms are evaluated from multiple aspects. We evaluate how well the task is finished from 1) average speed at each step (the closer to the targeted speed $8m/s$ the better); 2) average distance from the vehicle to the middle of the lane at each step (the smaller the better); 3) total number of steps that the vehicle outputs control actions in one run (the fewer the better). We evaluate the smoothness of the trajectories based on the variances of the speeds and distances in time series as well as the standard deviations of the speed and distance changes between consecutive steps, which can be regarded as an approximation of their derivatives over time. Smooth steering should induce low variances and standard deviations.} 
It is shown that before the traction-loss area is placed, the initial policy drives the vehicle at $8m/s$ on average and keeps the vehicle close to the middle of the lane. \add{Its low variances and standard deviations can be viewed as a baseline of the trajectory smoothness.} After the traction-loss event occurs, the initial policy still maintains the speed but the car slides out of the lane as indicated by the average distance. 
The initial policy and SC pair has the lowest average and lowest speed. \add{As a result, the total running steps increases. Its increased variances and the standard deviations signify that its steering gets less smooth.} In terms of policy repair, both Algorithm 1 and 2 are successful in finding safe policies. The policy repaired via Algorithm 1 behaves similar to the initial policy and SC pair -- the vehicle experiences significant speed changes and takes longer to finish the driving task.
The minimally deviating policy repaired via Algorithm 2 {behaves similarly} to the initial policy \add{in terms of maintaining the targeted speed, staying close to the middle of the lane while producing a smooth trajectory}. 
{In summary}, the repaired policy from Algorithm 2 outperforms the initial policy with SC \add{pair and the repaired policy from Algorithm 1} in {almost all metrics}. \add{We also observe that the average time of neural network inference is $0.0003s$ while the average time for SC to solve $\eqref{5}\sim\eqref{11}$ is $0.39s$}.

To further measure the impact of policy repair and evaluate the performance difference between a naive repair (using Algorithm 1) and a minimally deviating repair (using Algorithm 2), 
we remove the traction-loss area and execute both repaired policies in the original environment. 
It can be observed in Fig.\ref{fig19} that the naively repaired policy cuts inside the lane, since it learns (possibly due to overfitting) to steer inward in the states where traction loss is supposed to occur. 
In contrast, the policy repaired using Algorithm 2 manages to keep the car in the lane, as it learns to imitate the original policy. 
This thus further validates our approach of finding a minimally deviating repair.

\section{Conclusion}

We consider a Simplex architecture where a learning-based controller is paired with a backup safety controller for ensuring runtime safety. 
We show that this setup, while provides added safety assurance, can produce undesired outputs or cause significant performance degradation.
We propose to address this problem by
fine-tuning the learning-based controller using interventions from the safety controller, and addressing the issue of performance degradation via imitation learning. 
Our experiments indicate that our proposed approach is effective in achieving both safety and performance even when the dynamical model used by the safety controller is not exact.
In the future, we plan to consider other types of safety controllers and extend our techniques to end-to-end control.

\bibliographystyle{plain}
\bibliography{reference}
\end{document}